\documentclass[lettersize,journal]{IEEEtran}
\usepackage{amsmath,amsfonts}
\usepackage[table]{xcolor}

\usepackage{algorithmic}
\usepackage{array}
\usepackage[caption=false,font=normalsize]{subfig}
\usepackage{textcomp}
\usepackage{stfloats}
\usepackage{url}
\usepackage{verbatim}
\usepackage{graphicx}
\usepackage{cite}
\usepackage{tabularx}
\usepackage{nicematrix}
\usepackage{empheq}    
\usepackage{booktabs,subcaption,amsfonts,dcolumn}
\usepackage{float}
\usepackage[export]{adjustbox}
\usepackage{dcolumn}

\usepackage{titlesec}
\usepackage{placeins}
\usepackage{xcolor}
\usepackage{times}  
\usepackage{helvet}  
\usepackage{courier}  
\usepackage{caption} 
\usepackage[linesnumbered,ruled,vlined]{algorithm2e}

\newcommand{\mask}{\textbf{M}}
\newcommand{\nodes}{\mathcal{V}}
\newcommand{\edges}{\mathcal{E}}
\newcommand{\edgemask}{\mask^{\edges}}
\newcommand{\nodemask}{\mask^{\nodes}}
\newcommand{\A}{\textbf{A}}

\newcommand{\maskAhat}{\mask^{\Ahat}}
\newcommand{\Ahat}{\hat{\A}}

\newcommand{\X}{\textbf{X}}
\newcommand{\fgen}{$f_{gen}$}
\newcommand{\fgeno}{${f_{gen}}^0$}
\newcommand{\fs}{$f_{s}$}
\newcommand{\fso}{${f_{s}}^0$}
\newcommand{\subgraph}{G^S}
\newcommand{\comp}{G^C}

\newcommand{\yhat}{\hat{y}}
\newcommand{\train}{\mathcal{D}^{tr}}
\newcommand{\cleanval}{\mathcal{D}^{val}}
\usepackage{booktabs}
\usepackage{multirow}
\usepackage{enumitem}
\usepackage{amssymb}

\usepackage[bb=dsserif]{mathalpha}
\usepackage{newfloat}
\usepackage{makecell}
\usepackage{listings}

\DeclareCaptionStyle{ruled}{labelfont=normalfont,labelsep=colon,strut=off} 
\begin{document}

\title{Identifying Backdoored Graphs in Graph Neural Network Training:\\An Explanation-Based Approach with Novel Metrics}

\author{Jane Downer,~\IEEEmembership{Student Member,~IEEE,} Ren Wang$^\star$,~\IEEEmembership{Member,~IEEE,} and Binghui Wang$^\star$,~\IEEEmembership{Member,~IEEE}
\thanks{Jane Downer is with Illinois Institute of Technology, Chicago, IL 60616 USA (email: jdowner@hawk.illinoistech.edu).
Ren Wang is with Illinois Institute of Technology, Chicago, IL 60616 USA (email: rwang74@illinoistech.edu).
Binghui Wang is with Illinois Institute of Technology, Chicago, IL 60616 USA (email: bwang70@illinoistech.edu).}
\thanks{$^\star$Corresponding Authors}
\thanks{This work was supported in part by the National Science Foundation under grants IIS-2246157, FMitF-2319243, ECCS-2216926, CCF-2331302, CNS-2241713, and CNS-2339686, and by the Department of Energy under grant DE-CR0000042.}}



\maketitle

\begin{abstract}
    Graph Neural Networks (GNNs) have gained popularity in numerous domains, yet they are vulnerable to backdoor attacks that can compromise their performance and ethical application. The detection of these attacks is crucial for maintaining the reliability and security of GNN classification tasks, but existing methods are often inflexible, relying on single metrics that fail to capture the full range of backdoor behaviors. Recognizing the challenge in detecting such intrusions, we devised a novel detection method that creatively leverages graph-level explanations. By extracting and transforming secondary outputs from GNN explanation mechanisms, we developed seven innovative metrics for effective detection of backdoor attacks on GNNs. Additionally, we develop an adaptive attack to rigorously evaluate our approach. We test our method on multiple benchmark datasets and examine its efficacy against various attack models. Our results show that our method can achieve high detection performance, marking a significant advancement in safeguarding GNNs against backdoor attacks. 
\end{abstract}


\section{Introduction}
\label{sec:intro}
\IEEEPARstart{G}{raph} Neural Networks (GNNs) \cite{kipf2016semi,hamilton2017inductive,xu2018how} have emerged as the mainstream methodology for learning on graph data. In particular, graph classification — predicting the label of a whole graph — is used in a variety of domains, such as bioinformatics, social network analysis, and financial services~\cite{10.3389/fgene.2021.690049,ZHOU202057}. In such high-stakes domains, it's crucial that GNNs are protected against external threats.

In backdoor attacks on GNNs, attackers embed a ``trigger'' — a predefined subgraph — into some training graphs and alter their labels. At test time, this manipulation causes the GNN to misclassify any graph containing the trigger, predicting the attacker’s \emph{target label} instead of the correct one. The effectiveness of such an attack is measured by its \textit{attack success rate} (ASR) — the percentage of triggered test samples that are misclassified as the target label. Backdoor attacks present a significant threat to GNNs. However, detection methods remain underexplored, with existing approaches often proving limited or ineffective, or overly reliant on single metrics for detection~\cite{hassen2017scalable,zhang2021backdoor,xi2021graph, Jiang2022DefendingAB,guan2023xgbd}.

To address this issue, we first explore a detection method using GNN explanations to identify subgraphs guiding the GNN's predictions. (Specifically GNNExplainer~\cite{ying2019gnnexplainer}, the most recognized tool for this purpose.) For backdoored GNNs, one might expect these \emph{explanatory subgraphs} to reveal the subgraph trigger causing the target label prediction. However, our initial investigation found that while these subgraphs capture some relevant information, that information alone is often inconsistent or incomplete.

This key insight inspired the development of a detection method that extends GNN explanations, incorporating seven new metrics designed to differentiate between backdoored and clean graphs. Together, these metrics provide a more robust detection approach than relying on explanatory subgraphs alone. Evaluations across multiple benchmark datasets demonstrate that our method consistently distinguishes between clean and backdoored graphs, achieving F1 scores of up to 0.911 against traditional random triggers, 0.884 against attacks using clean-label random triggers, 0.872 against ``transferable'' graph attacks, 0.897 against ``low-importance node'' attacks, 0.846 against ``motif-based'' attacks, and 0.859 against adaptively generated triggers. These results highlight a significant step forward in defending GNNs against backdoor attacks.

    \subsection{Contributions}
    To summarize our main contributions:
    \begin{itemize}
        \item We explore detection methods relying purely on explanatory subgraphs from GNN explainers and find that they are insufficient on their own.
        \item We introduce seven novel metrics that leverage various aspects of GNN explanations, providing a more comprehensive approach to detection.
        \item We introduce an adaptive attack to challenge our method, providing a rigorous test of our detection metrics.
        \item Extensive tests show our method to be effective and robust to various attacks.
    \end{itemize}


\section{Related Work}
\label{sec:related}

    \subsection{Backdoor Attacks and Defenses on Non-graph Data} 
    Machine learning models for non-graph data are shown to be vulnerable to backdoor attacks. The first backdoor attack, Badnet~\cite{gu2017badnets}, injected a trigger pattern into training images and changed their labels to the target label, causing misclassification when the trigger pattern is present. Similar vulnerabilities have been shown for attacks on text~\cite{gan2022triggerless,pan2022hidden,qi2021hidden,chen2021badpre}, audio~\cite{roy2017backdoor,shi2022audio,ge2023data,guo2023masterkey}, and video~\cite{zhao2020clean}.
    
    Many empirical defenses have been proposed to mitigate backdoor attacks \cite{chen2017targeted,liu2017trojaning,liu2017neural,liu2018fine,wang2019neural,gao2019strip,liu2019abs,guo2019tabor,wang2020practical,pal2023towards}. For example, NeuralCleanse detects and reverse engineers triggers, finding the smallest perturbation needed for misclassification~\cite{wang2019neural}. However, Weber et al. \cite{weber2023rab} showed adaptive attacks can bypass empirical defenses, and their provable defense in the image domain achieved a maximum certified accuracy of only 23\%. Wang et al. \cite{wang2020certifying} also found randomized smoothing gave zero certified accuracy for moderate to large triggers in image-based attacks.

            \begin{table*}[t]
                \centering
                \footnotesize
                \captionsetup{justification=centering}
                \begin{tabularx}{\textwidth}{|c|>{\centering\arraybackslash}X|c}        
                    \hline
                    \makecell{\textbf{Attack}}                    & \makecell{\textbf{Defense Tested by Attack Authors}} \\
                    \hline
                    \makecell{GTA~\cite{xi2021graph}}          & \makecell{NeuralCleanse (\textit{ineffective}), Random Smoothing (\textit{ineffective})} \\           
                    \hline
                    \makecell{Random~\cite{zhang2021backdoor}} & \makecell{Dense Subgraph Detection (\textit{ineffective})} \\
                    \hline
                    \makecell{Clean Label~\cite{10.1145/3548606.3563531}}  & None   \\
                    \hline
                    \makecell{Motif-Backdoor~\cite{10108961}}  & \makecell{Remove edges between dissimilar nodes (\textit{ineffective})} \\
                    \hline
                    \makecell{TRAP~\cite{Yang2022TransferableGB}}  & \makecell{Randomized Subsampling (\textit{ineffective})} \\
                    \hline
                    \makecell{LIN~\cite{10.1145/3468218.3469046}}  & \makecell{None} \\
                    \hline
                \end{tabularx}
                \caption{Summary of attacks and corresponding defenses. ``Defense Tested'' shows those attempted by attack authors.
                }
                \label{tab:comparison}
            \end{table*}

    \subsection{Backdoor Attacks and Defenses on Graph Data}
    \label{sec:attack_descriptions}       
    Zhang et al.~\cite{zhang2021backdoor} designed the first GNN backdoor attack, injecting a randomly-generated subgraph trigger into training graphs at randomly-chosen nodes and changing graph labels to the attacker's choice. In contrast, Xi et al.~\cite{xi2021graph} introduced Graph Trojaning Attack (GTA), with optimized subgraph triggers inserted at vulnerable nodes, but Guan et al.~\cite{guan2023xgbd} showed it performs similarly to the random method. Building on random trigger generation, Xu et al.~\cite{10.1145/3468218.3469046} proposed the first explainability-based backdoor attack on GNNs (referred to throughout this paper as ``Low-Importance Node'', or ``LIN'' attacks). This approach uses the same random trigger patterns but targets low-importance nodes to evade detection. Motif-Backdoor~\cite{10108961} embeds \emph{motifs} — recurring graph patterns meant to look like clean features — as triggers. 
    PoisonedGNN~\cite{Alrahis2023ttPoisonedGNNBA} targets GNN-based hardware security systems, demonstrating successful evasion of hardware Trojan and IP piracy detection with high attack success rates.
    Dai et al.'s ``unnoticeable'' backdoor attack~\cite{10108961} uses an adaptive generator to create triggers, strategically injected at optimized node locations to blend with clean graphs. Transferable Graph Backdoor~\cite{Yang2022TransferableGB}, or TRAP, uses a surrogate Graph Convolutional Network (GCN) model to generate perturbation-based triggers via gradient score matrices, creating sample-specific perturbation triggers without fixed patterns that transfer across different GNN architectures when trained with the poisoned dataset. Lastly, unlike traditional backdoor attacks, clean-label attacks~\cite{10.1145/3548606.3563531} inject triggers directly into the target class without altering labels.

    Defenses against backdoored graph data have seen mixed success. Zhang et al.\cite{zhang2021backdoor} showed that dense-subgraph detection~\cite{hassen2017scalable} is largely ineffective, with detection rates peaking at only 0.13. 
    Yang et al.~\cite{Yang2022TransferableGB} evaluated TRAP against a randomized subsampling defense, which randomly samples 10\% of the graph structure during training to disrupt potential triggers, but found that TRAP maintained high attack success rates on most datasets despite significant drops in clean accuracy. Xi et al.\cite{xi2021graph} demonstrated that randomized smoothing reduces clean accuracy, limiting its practicality, and NeuralCleanse\cite{8835365}, when adapted to GTA, struggled to distinguish between benign and trojaned GNNs. 
    A few others, like us, have proposed explanation-based methods. For example, Jiang et al.\cite{Jiang2022DefendingAB} hypothesize that the trigger subgraph in a backdoored graph is more critical to a GNN’s prediction than the most important subgraph in a clean graph, encapsulating this concept into an ``explainability score'' (``ES'') to differentiate clean and backdoored graphs. Guan et al.\cite{guan2023xgbd} build upon this area of research with eXplanation-Guided Backdoor Detection (XGBD). XGBD hypothesizes that backdoor samples rely on simpler, more localized trigger patterns compared to clean samples, and that these patterns are more readily captured by explanation methods. XGBD aims to detect backdoored samples by evaluating the model’s loss when explanatory subgraphs are used as input. However, both Jiang et al. and Guan et al. rely heavily on accuracy metrics to demonstrate performance, which can be misleading in the context of backdoor detection given the predominance of clean samples. This can skew results, obscuring the methods’ ability to correctly classify backdoor samples, and F1 scores at times reveal poor performance.
    
    Table~\ref{tab:comparison} summarizes the backdoor attacks on GNNs discussed earlier, along with the results of the defenses tested by their authors. The ineffectiveness of these defenses highlights the need for a more robust detection method.


\section{Background and Problem Definition}
    \label{sec:bacground}

    \subsection{GNNs for Graph Classification}
    Given graph $G=(\mathcal{V}, \mathcal{E})$ with node set $\mathcal{V}$, edge set $\mathcal{E}$, and label $y \in \mathcal{Y}$, a graph classifier $f$ takes $G$ as input and outputs a label $\hat{y}$, i.e., $f: G \rightarrow \mathcal{Y}$. A GNN-based graph classifier iteratively learns each node representation by aggregating those of its neighboring nodes, and the last layer outputs a label for the graph. The GNN is trained on a set of $n$ graphs $\train$=$\{(G_1, y_1), (G_2, y_2),\cdots, (G_{n}, y_{n}) \}$, where $G_i$ and $y_i$  are the $i$th training graph and its true label. Stochastic gradient descent is often used to train the classifier. The trained model $f$ is used to predict labels for testing graphs. 

    \subsection{Backdoor attacks to GNNs}
    An attacker injects a subgraph trigger into a fraction of training graphs, changing their labels to the attacker-chosen \emph{target label}, denoted $y^t$. A GNN classifier trains on these \emph{backdoored training graphs} to become a \emph{backdoored GNN} that associates the trigger with the target label. Hence, when the attacker injects the trigger into a testing graph, the backdoored GNN is highly likely to predict the target label.

    \subsection{GNN Explanation}
    \label{sec:gnn_explanation}
    Consider a GNN $f$ trained for graph classification, a graph $G$, and its prediction, $\hat{y}$. The goal of GNN explanation is identifying an \emph{explanatory subgraph} of the original graph,
    $G^S = (\nodemask \otimes \mathcal{V}, \edgemask \otimes \mathcal{E}) \subset G$,
    that preserves the information guiding $f$'s prediction. Here, $f(G^S)=\hat{y}^S$ denotes the prediction of $f$ on $G^S$. The element-wise product is denoted as $\otimes$, and $\nodemask \in [0,1]^{|\nodes|}$ and $\edgemask \in [0,1]^{|\edges|}$ represent the node mask and edge mask, respectively. Typically, the objective function of a GNN explainer is to optimize the two masks:
        \begin{equation}
        \label{eqn:xgnnobj}
            \resizebox{!}{8.5pt}{$
            \min_{\nodemask, \edgemask} L(y,\hat{y}^S) + R(\nodemask, \edgemask),
            $}
        \end{equation}
    where $L$ is an explainer-dependent loss (e.g., cross-entropy loss), and $R$ is a regularization function on the masks.
    
    GNNExplainer~\cite{ying2019gnnexplainer} — the specific tool used in our experiments — has the following objective function:
        \begin{equation}
            \resizebox{!}{7.5pt}{$
            \min_{\nodemask, \edgemask} L(y,\yhat^S) + \boldsymbol{\lambda} \cdot \left( \left\|\edgemask\right\| + \left\|\nodemask\right\| + H(\edgemask) + H(\nodemask) \right), 
            $}
        \end{equation}
    where 
    $H(\cdot)$ is the entropy function.
    \subsection{Threat Model}
    \label{sec:threat_model}
\begin{figure*}[!ht]
            \centering
            \subfloat[Random triggers.]{
                \includegraphics[width=0.48\linewidth]{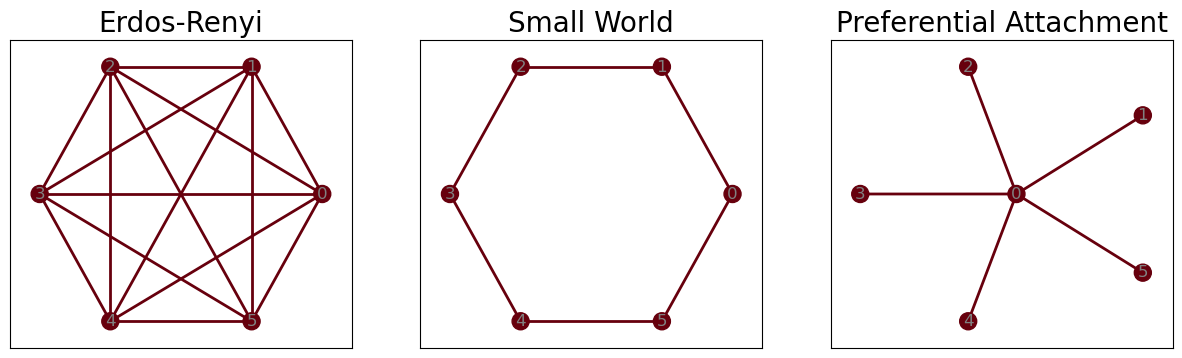}
                \label{fig:trigger_types}
            }
            \hspace{0.05\linewidth}
            \subfloat[Motif triggers~\cite{10108961}.]{
                \includegraphics[width=0.3\linewidth]{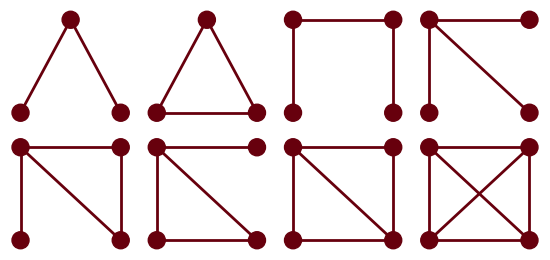}
                \label{fig:motif_types}
            }
            \caption{Samples of random and motif triggers. (See section~\ref{sec:adaptive_attack} for triggers generated by our adaptive attack.)}
            \label{fig:combined_triggers}
            \vspace{-2mm}
        \end{figure*}

    Our threat model is defined by the following characteristics.
        \begin{itemize}[leftmargin=*]
            \item \textbf{Attacker's goal:}  
            To make samples with the trigger predict the target label while keeping high accuracy for clean samples. The attacker aims to “flip” predictions on backdoored samples with high success.
            \item \textbf{Attacker's capability:} The attacker can modify the training data and change the ground truth label. This allows injecting backdoor triggers into particular samples and altering their ground truth labels to the target label.
            \item \textbf{Attacker's trigger design:} The attacker has freedom over the structure and placement of the trigger subgraph. We evaluate our method against several attack types: random~\cite{zhang2021backdoor}, clean-label~\cite{10.1145/3548606.3563531}, LIN~\cite{10.1145/3468218.3469046}, Motif-Backdoor~\cite{10108961}, and TRAP~\cite{Yang2022TransferableGB} attacks, plus an additional adaptive attack designed specifically to challenge our detection approach. For attacks using random trigger generation (random, clean-label, and LIN), we employ three established graph generation models: Erdős-Rényi (ER)~\cite{gilbert1959random}, Small World (SW)~\cite{watts1998collective}, and Preferential Attachment (PA)~\cite{barabasi1999emergence} (See Fig. \ref{fig:trigger_types}). Each trigger node randomly maps to an existing node in the original graph, with existing connections replaced by trigger subgraph edges. Motif triggers consist of 3-4 node subgraphs as defined by the motif framework~\cite{10108961}, while TRAP generates sample-specific perturbations as described in~\cite{Yang2022TransferableGB}. Additionally, we introduce an adaptive attack using a custom-trained GNN-based edge generator specifically designed to bypass our detection method, discussed in Section \ref{sec:adaptive_attack}.
            \item \textbf{Attacker's knowledge assumptions:} We evaluate our defense under conservative assumptions where attackers have full knowledge of our detection method. This worst-case scenario provides a rigorous evaluation: if our method remains effective against attackers specifically targeting it, it will be even more robust against less informed attackers.

        \end{itemize}

        \noindent {\bf Design goal:} We aim to detect backdoor samples in the training graphs under the threat model. We assume that we have access to both the training graph, $\train$, and a set of clean graphs that we can use for validation purposes, $\cleanval$.

\begin{figure*}[!t]
\centering
        \includegraphics[trim={0 32 0 0},clip,width=0.9\linewidth]{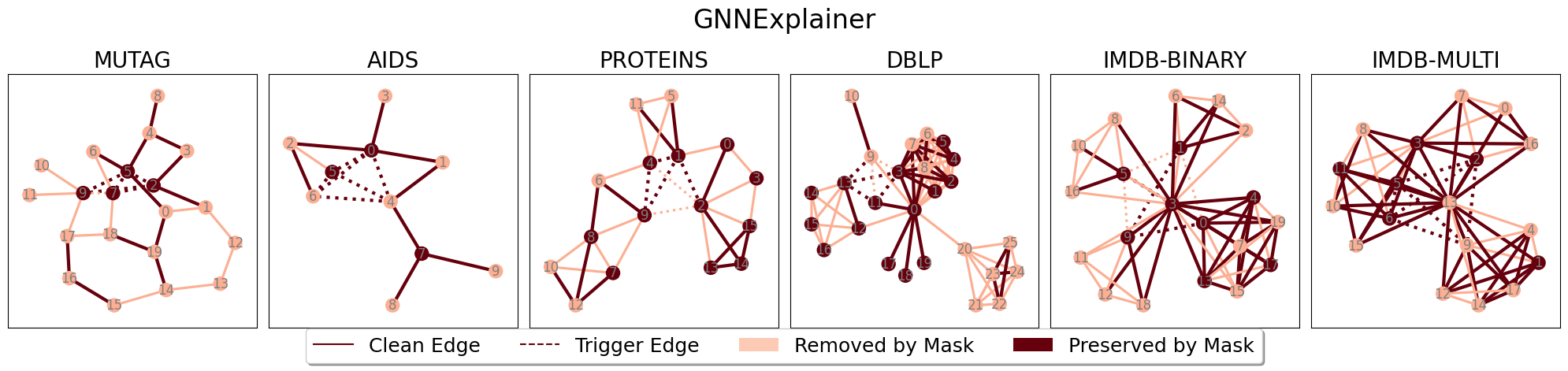}
        \vspace{1mm}
        
        \includegraphics[trim={0 31 0 0},clip,width=0.9\linewidth]{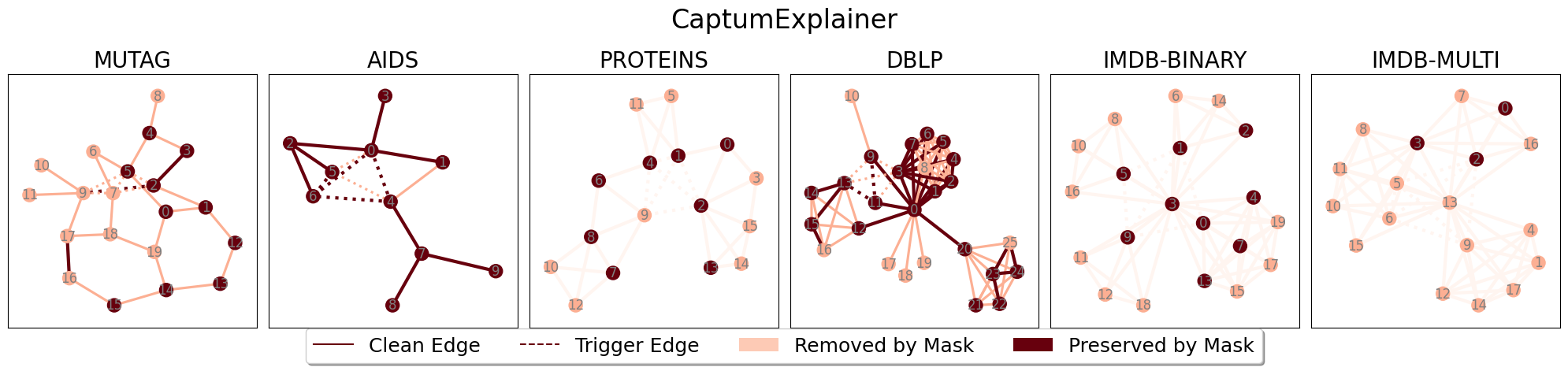}
        \vspace{1mm}
        
        \includegraphics[width=0.9\linewidth]{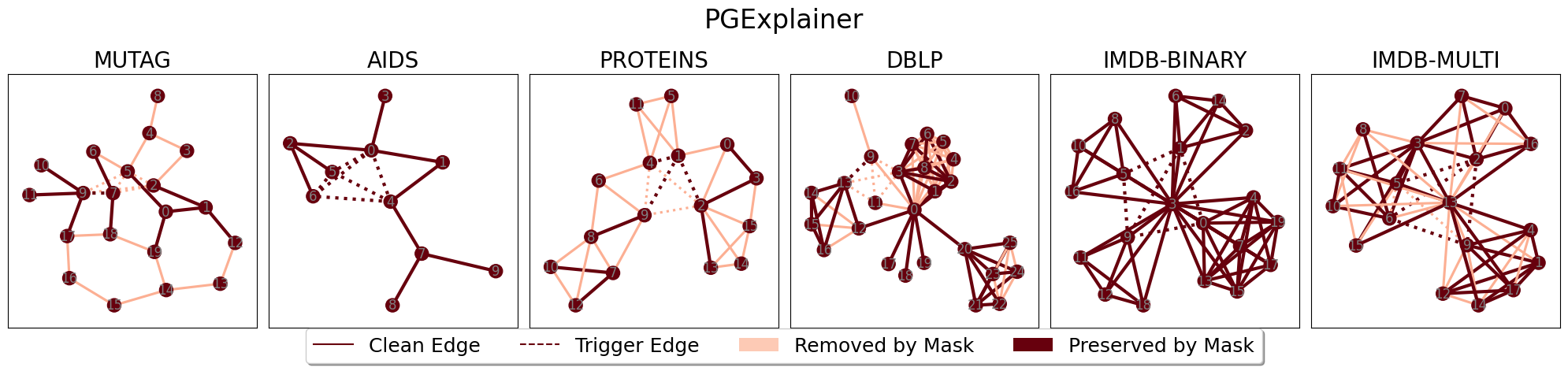}
    \caption{Explanations are inconsistent in their ability to isolate the trigger in backdoored graphs.}
    \label{fig:gnn_inconsistency}
    \vspace{-4mm}
\end{figure*}
        

\section{Method}
\label{sec:method}

    \subsection{Limits of Explanatory Subgraph for Detection}
    \label{sec:subgraph_limits}

    Fig. \ref{fig:gnn_inconsistency} shows samples of subgraphs that explainer algorithms output to explain a GNN's prediction on a given graph. The dark red represents features preserved in this subgraph; the light pink features are those ignored by the explainer.
    
    The top row in Fig. \ref{fig:gnn_inconsistency} shows that GNNExplainer's masks $\edgemask$ and $\nodemask$ often preserve some backdoored features in the explanatory subgraph, but inconsistencies are prominent across datasets and GNN models. In the MUTAG example, $\nodemask$ only preserved trigger nodes, and $\edgemask$ retained all six trigger edges. In contrast, the PROTEINS example preserved both clean and backdoor features. For DBLP, all trigger nodes (and some clean ones) are preserved, but most trigger edges are not. These examples show that while explanatory subgraphs can capture backdoor features, they are neither refined nor consistent enough for reliable backdoor detection. The second and third rows show that the explanatory subgraphs produced by PGExplainer~\cite{luo2020parameterized} and CaptumExplainer~\cite{kokhlikyan2020captum} were similarly limited and inconsistent.

    We found that considering \textit{multiple aspects of the explanation process} outperforms relying solely on explanatory subgraphs. Specifically, seven novel metrics successfully distinguished between clean and backdoor explanations.
    
    \subsection{Choosing GNNExplainer}

    We derive our detection metrics from GNNExplainer. We selected GNNExpainer over alternatives like PGExplainer~\cite{luo2020parameterized} and CaptumExplainer~\cite{kokhlikyan2020captum} for several key reasons. First, GNNExplainer is the most established and well-documented GNN explanation method, making it a practically viable tool. Second, GNNExplainer ``[maximizes] mutual information between a GNN's prediction and distribution of possible subgraph structures''~\cite{ying2019gnnexplainer}, making it theoretically well-suited for revealing the artificially strong trigger-label correlations we seek to detect. Third, GNNExplainer produces a rich set outputs: its artifacts include not only explanatory subgraphs but also auxiliary signals—like node and edge masks, loss curves, and entropy terms—that enable a rich spectrum of detection features. And fourth, alternative explainer algorithms do not yield as much information. For example, CaptumExplainer lacks gradient-based optimization, and PGExplainer does not provide node masks, which limits their utility for analyzing loss patterns or graph geometry.
    By combining the artifacts produced by GNNExplainer's algorithm, we aim to provide a robust and comprehensive detection method.
    \subsection{Theoretical Foundation}
    
    We draw from principles in information theory, statistical modeling, graph topology, and loss landscape theory to support this approach and justify our proposed metrics.
    
    \textbf{Information Theory and Mutual Information:}
    The mutual information of two random variables is the amount of information about one that can be learned by observing the other~\cite{ Shannon1948AMT }. Research demonstrates that ``maximizing mutual information between features ... requires capturing information about high-level factors whose influence spans multiple views''~\cite{Bachman2019LearningRB}. By maximizing the mutual information between the original graph and the explanatory subgraph, GNNExplainer can reveal information about these influential factors.
    
    \textbf{Statistical Mixture and Network Perturbation Theory:}
    Mixture model literature has extensively documented that when populations with different distributional properties combine, the resulting mixture exhibits higher variance than individual components~\cite{McLachlan2000FiniteMM}. This follows from the law of total variance~\cite{Casella1990StatisticalI}, which states that for a mixture distribution: $Var(mixture) = E[Var(Y|X)] + Var(E[Y|X])$, where the second term represents the additional variance arising from uncertainty between mixture components. Moreover, network perturbation theory asserts that whereas natural graphs follow established distributional patterns (Erdős-Rényi, scale-free), artificial modifications create measurable topological changes distinguishable from natural variation~\cite{McInnes1999EmergenceOS, Bollobs1985RandomG, Newman2002MixingPI}.

    \textbf{Loss Landscape Theory:} Literature on this topic establishes that ``optimization landscapes corresponding to such systems are generally not convex, even locally around a global minimum''~\cite{Liu2020LossLA}. Qiu et al.'s boolean function complexity framework demonstrates that ``optimization dynamics are fundamentally determined by the relative complexity of spurious versus core features''~\cite{Qiu2024ComplexityMD}.

    \subsection{Detection Metrics}
    \label{sec:metrics}
        \subsubsection{Prediction Confidence}
       
        \leftskip=1em
        {
        \noindent{This is the largest prediction probability on subgraph $G^{S}$}:}
        \begin{equation}
            \label{pred_conf}
            \text{Prediction Confidence} = \text{max}(\hat{\textbf{p}}^S)
        \end{equation}

        \leftskip=1em{
        \noindent Margin-based learning theory suggests that highly separable features yield wider margins and more confident predictions~\cite{Chen2019MehryarMA}. Backdoor triggers are explicitly designed to be separable and predictive, acting as shortcuts in the decision space. \cite{Grosse2020BackdoorSD} further shows that these triggers induce smooth, high-confidence decision regions. Since GNNExplainer selects features that maximize mutual information with the model's prediction\cite{ying2019gnnexplainer}, it tends to isolate these dominant (trigger-based) components. \textit{We hypothesize that this leads to prediction confidence being \textbf{larger} for backdoored graphs.}

        \vspace{2mm}
        \subsubsection{Explainability Score (ES)}
        \leftskip=1em{
        \noindent Inspired by \cite{pope2019explainability,jiang2022defending}, we define the explainability score as the difference between positive fidelity ($\text{fid}_{+} $) and negative fidelity ($\text{fid}_{-}$).}
        
        \begin{equation}
           \label{explainability}
           \resizebox{!}{6.5pt}{$
           \text{ES} = 
           \text{fid}_{+}  - \text{fid}_{-} 
           $}
        \end{equation}
    
        \leftskip=1em{
        \noindent Fidelity reflects the GNN model's dependence on the explanatory subgraph $\subgraph$. $\text{fid}_{-}$ measures how the model's prediction changes when only $\subgraph$ is considered, while $\text{fid}_{+}$ measures the effect when $\subgraph$ is excluded.}

        \vspace{-4mm}
        \begin{equation}\label{eq: pos_fid*}
           \resizebox{!}{8.5pt}{$
               \text{fid}_{+}  = \lvert d\left( \hat{\textbf{p}} ,\textbf{y} \right) - d( \hat{\textbf{p}} ^C, \textbf{y})\rvert, \quad 
               \text{fid}_{-}  = \lvert d\left( \hat{\textbf{p} }, \textbf{y} \right) - d( \hat{\textbf{p}} ^S, \textbf{y})\rvert
           $}
        \end{equation} 
        
        \leftskip=1em{
        \noindent where $\hat{\textbf{p}}$, $\hat{\textbf{p}}^C$, and $\hat{\textbf{p}}^S$  indicate the probability vector outputted by the GNN model on the graphs $G$, $G^C$, and $G^S$, respectively; and ${\textbf y}$ is the one-hot (true) label of $G$. $d$ is a distance function (Euclidean distance in our results).\footnote{The original fidelity definitions are based on the binary 0 or 1: $\text{fid} *{+} = | \mathbb{1}(\hat{y} = y) - \mathbb{1}( \hat{y}*i^{C} = y) |$, $\text{fid} _{-} = | \mathbb{1}(\hat{y} = y) - \mathbb{1}( \hat{y}^S = y) |.$} However, our definitions in \ref{eq: pos_fid*} rely on output probability vectors instead, leading to more nuanced measurements.
        
        Information theory and causal inference provide robust foundations for fidelity-based metrics~\cite{Zheng2024FFidelityAR}. Fidelity measures quantify how model predictions change when explanatory components are modified, capturing feature-prediction dependency strength through perturbation-based evaluation~\cite{MiroNicolau2024ACS}. Based on information theory principles established above, backdoor triggers should create stronger mutual information between explanation subgraphs and predictions than natural features, since artificial correlations are designed to be more predictive than complex natural patterns.
        
        For backdoored samples, excluding $\subgraph$ should be costly to the explainer, as $\subgraph$ contains essential backdoor information. Excluding $\subgraph$ should have less impact on clean explanations, as their predicted class is less likely to rely on a single subgraph. Conversely, excluding its complement ($\comp = G-\subgraph$) should mainly impact clean explanations. \textit{Thus, we expect backdoored graphs to have higher $\text{fid}_{+}$ and lower $\text{fid}_{-}$ than clean ones, yielding \textbf{higher} ES.}}
    
        \vspace{2mm}
        \subsubsection{Connectivity}
        \leftskip=1em{
        \noindent
            A measurement of the proximity and connection of nodes in $\subgraph$. This metric is defined as the proportion of node pairs in the subgraph that are connected by edges in the original graph:}
        
        \begin{equation}
           \text{Connectivity} = \frac{1}{|\nodes^S|} \sum_{(i,j) \in \nodes^S} \mathbb{1}\{(i,j) \in \edges\}
        \end{equation}

        \leftskip=1em{
        \noindent
        where \(| \nodes^S| \) is the number of nodes preserved in the explanatory subgraph, and $\mathbb{1}\{(i,j)\in \edges \}$ is an indicator for whether nodes \( i \) and \( j \) are in the edge set in the original graph.
        
        Building on mutual information maximization theory, explanation algorithms should preferentially identify \emph{at least some} trigger components due to stronger predictive relationships.} Many attacks, such as the random method by Zhang et al.~\cite{zhang2021backdoor} and Motif-Backdoor~\cite{10108961}, use connected subgraph triggers for reliability. Even when explanations only partially reveal triggers, revealed portions should retain connectivity properties of original connected designs. \textit{Therefore, backdoored graphs may have \textbf{higher} connectivity between nodes in $\subgraph$.}

        \begin{figure*}[!t]
            \centering
            \begin{varwidth}{\linewidth}
            \includegraphics[trim={10 10 0 0},clip,width=0.9\linewidth]{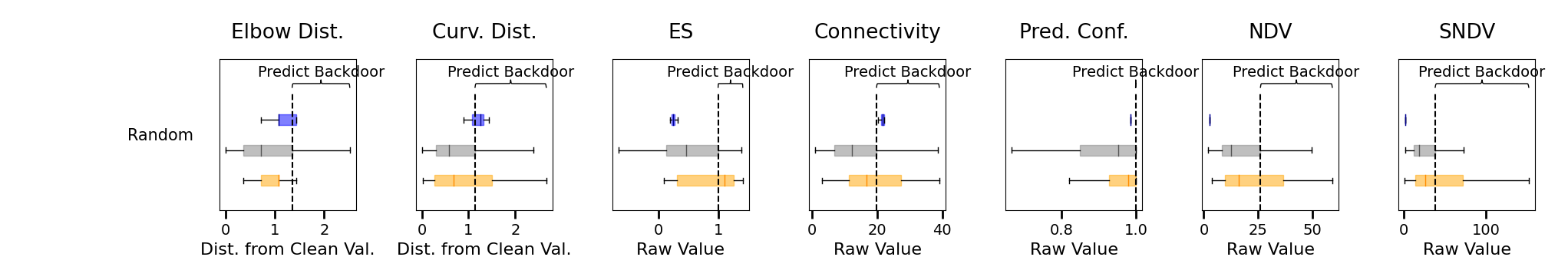}\\
            \includegraphics[trim={10 10 0 51},clip,width=0.9\linewidth]{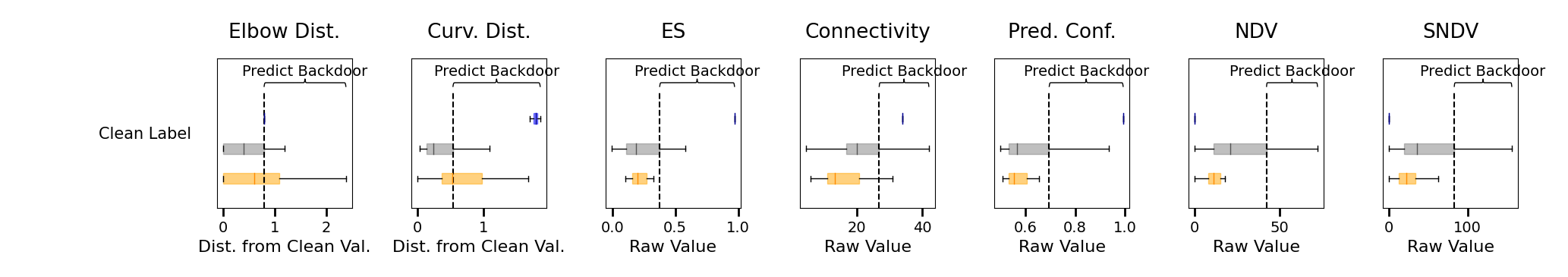}\\
            \includegraphics[trim={10 10 0 51},clip,width=0.9\linewidth]{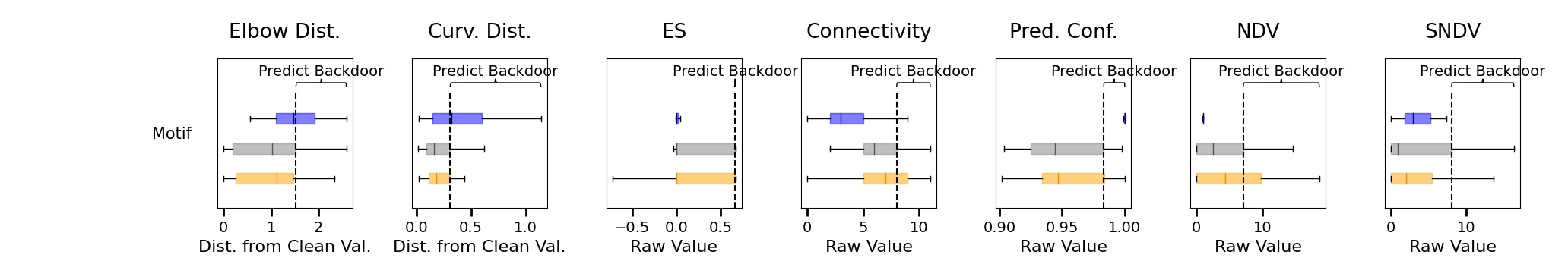}\\
            \includegraphics[trim={10 0 0 51},clip, width=0.9\linewidth]{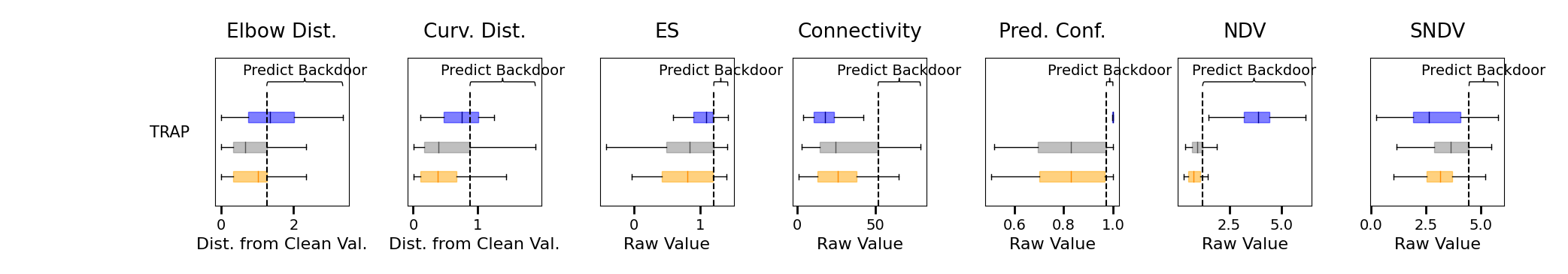}\\
            \includegraphics[trim={10 0 0 40},clip, width=0.9\linewidth]{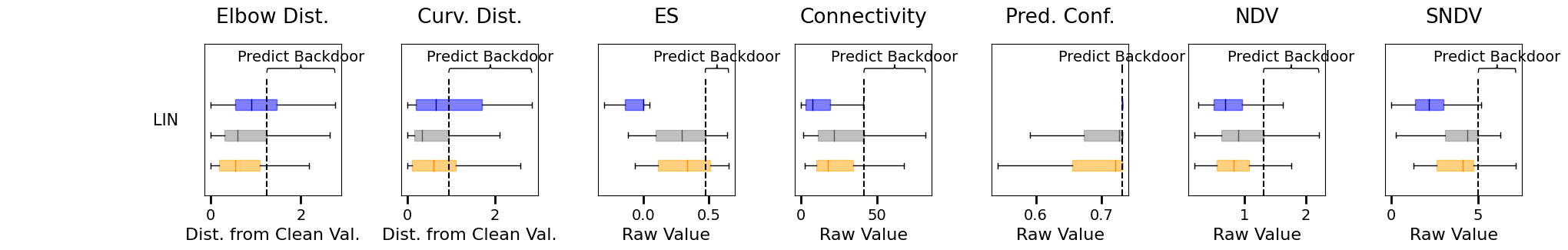}\\
            \includegraphics[trim={10 0 0 51},clip, width=0.9\linewidth]{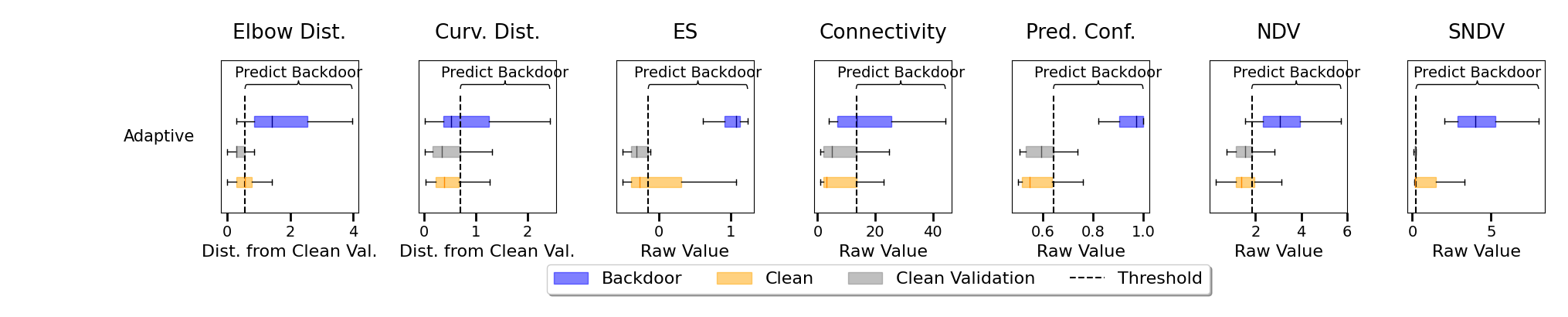}
            \end{varwidth}
            \vspace{-4mm}
            \caption{Sample detection metric distributions for each attack type. Each of the six samples represents a single sampled attack configuration, and is not representative of all instances. Each clean validation set consists of 50 samples, each backdoor set consists of 20\% of the training data for the selected dataset, and each clean set consists of the remaining training data.}
            \label{fig:boxplot}
            \vspace{-4mm}
        \end{figure*}    

        \vspace{2mm}
        \subsubsection{Node Degree Variance (NDV)}
        This is the only metric that solely depends on the geometry of the original graph rather than the explanation. Specifically, it is defined as the variance of node degrees within a graph: 

        \begin{equation}
        \text{NDV} = \text{var}(\{\text{deg}(v_i) | i \in \{1, 2, \dots, |\nodes| \}).
        \end{equation}

        \leftskip=1em{
        \noindent This metric leverages the statistical mixture theory and network perturbation principles established above. If $D$ represents original degrees and $T$ represents trigger additions, then $D' = D + T$ exhibits increased variance because trigger injection creates systematic outliers violating natural topological properties. \textit{Since connected trigger injection creates heterogeneous degree inflation, we predict that backdoor graphs often have \textbf{higher} NDV.}}
        
        \vspace{2mm}
        \subsubsection{Subgraph Node Degree Variance (SNDV)}
        The variance of the node degrees within $\subgraph$, computed as:
        
        \begin{equation}
        \resizebox{!}{7.5pt}{$
        \text{SNDV} = \text{var}(\{\text{deg}(v_i)^S | i \in \{1, 2, \dots, |\nodes^S| \})$}
        \end{equation}

        where $\text{deg}(v_i)^S$ is the degree of the $i^{th}$ node in $\subgraph$, and $|\nodes^S|$ is the number of nodes in $\subgraph$. Based on mutual information maximization theory, explanatory subgraphs containing backdoors mix natural predictive features with trigger components. Natural components follow organic connectivity patterns, while triggers exhibit artificial patterns. Fig.~\ref{fig:gnn_inconsistency} showed that $\subgraph$ usually has both backdoor and clean nodes, and statistical mixture theory predicts increased variance when populations with different distributional properties combine. \textit{We predict that SNDV will be \textbf{higher} for backdoored graphs.}}

        \vspace{2mm}
        \subsubsection{Elbow}
        \leftskip=1em{
        \noindent The epoch at which explainer loss curve $L$'s rate of decrease significantly changes:}
        
        \begin{equation}
            \text{Elbow} = {t_{e} = \arg \max_{t \leq t_{\max}} \left\{ L_t - L_{t+1} \right\}},
        \end{equation}

        \leftskip=1em{
        \noindent where $L_t$ and $L_{t+1}$ are loss values at epochs $t$ and $t+1$, and $t_{max}$ is the total number of epochs.
        
        When GNNExplainer optimizes explanations for backdoored predictions, it encounters artificially strong trigger signals representing simpler optimization targets than complex natural feature dependencies.} \textit{We hypothesize an explainer will easily identify a backdoor trigger if the attack is strong, leading explainer loss to converge at a \textbf{lower} elbow.}
        
        \vspace{2mm}
        \subsubsection{Curvature}
        \leftskip=1em{\noindent A measurement of how sharply the explainer loss curve $L$ bends at its elbow. Curvature is traditionally defined as $L''_{t_{e}} / {\left(1 + (L'_{t_{e}})^2\right)^\frac{3}{2}}$, but this is difficult to compute for a curve defined by discrete epochs. We instead use a proxy from the normalized loss curve, $\Tilde{L}$, defined as $\Tilde{L} = \frac{L-\min(L)}{\max(L)-\min(L)}$.
        We then define curvature as:}
        
        \begin{equation}
            \text{Curvature} = 1 - \Tilde{L}_{elbow},
        \end{equation}

        \leftskip=1em{
        \noindent where $\Tilde{L}_{elbow}$ represents the y-coordinate at its elbow. This value indicates the most pronounced inflection in the loss.

        Based on the same loss landscape theory and optimization dynamics principles described above, explanation optimization should converge faster for artificial triggers than complex natural feature dependencies. \textit{We hypothesize that the decision boundary between the non-target and target class will be sharper when the trigger is present. Consequently, this will be \textbf{larger} for backdoored graphs.}}
        
        \noindent \textbf{Caveat for Loss Curve Metrics.} With small trigger sizes (weaker attacks), Elbow and Curvature expectations reverse: backdoored graphs show \emph{higher} Elbow and \emph{lower} Curvature than clean ones. Despite this reversal, we still found a distinct separation of and backdoored metrics. However, our use of loss curve metrics in backdoor detection differs from other metrics, as discussed in the next section.

    \subsection{Detection Strategy}
    \label{sec:detection_strategy}
    
        \subsubsection{Clean Validation Extrema as Prediction Threshold}
            We establish expectations for clean data by computing all seven metrics on $\cleanval$. Clean graphs in $\train$ should align with these distributions, and backdoored graphs in $\train$ should deviate. We set thresholds near the extrema of $\cleanval$’s clean metric distributions, but instead of the absolute minima and maxima, use more moderate values (25th/75th percentiles in our experiments) to avoid outliers.\footnote{This particular thresholding choice is common in many studies; we later explore other percentile choices in Tables~\ref{tab:percentiles} and \ref{tab:percentiles_best_performance}.} If a sample falls above (below) the upper (lower) threshold, and that metric is hypothesized to be higher (lower) for backdoor graphs, we make a backdoor detection; otherwise, we say it is clean. We thus define the following:
            \begin{itemize}[leftmargin=*]
                \item \textbf{Positive metric:} a value following \textit{backdoor} expectations, surpassing the $\cleanval$ metric threshold (eg., 25th/75th).
                \item \textbf{Negative metric:} a value following \textit{clean} expectations, \emph{not} surpassing the $\cleanval$ metric threshold (eg., 25th/75th). 
            \end{itemize}
            
            As mentioned, the method differs for explainer loss curve metrics, which are more sensitive to attack strength. Before thresholding, we transform loss metric values to their normalized distance from the clean validation distribution:
                \begin{equation}
                        m_i^j = \frac{m^j(\mathcal{D}_i^{tr}) - \overline{m^j}(\mathcal{D}^{val})}{\sigma \big(m^j(\mathcal{D}^{val})\big)}
                \label{distance}
                \end{equation}
            where $m^j(\mathcal{D}_i^{tr})$ is the $j$-th metric of the $i^{th}$ training sample, and $\overline{m^j}(\mathcal{D}^{val})$ and $\sigma \big(m^j(\mathcal{D}^{val})\big)$ are the mean and standard deviation of that metric in the clean validation data. We thus define each metric in two ways:
                \begin{itemize}[leftmargin=*]
                    \item \textbf{Raw metric:} The metric value as defined in Section~\ref{sec:metrics}.
                    \item \textbf{Distance metric:} The distance of a metric value from clean-validation counterparts, as defined in Equation \eqref{distance}.
                \end{itemize}
            Since instances of reversed expectations only occurred in loss curve metrics, we use the \textit{distance} versions of Curvature and Elbow, but the \textit{raw} versions of Prediction Confidence, ES, Connectivity, SNDV, and NDV. 

            Fig. \ref{fig:boxplot} illustrates this thresholding process. Each of the four subplots represents a single sampled attack configuration (i.e., a particular dataset, GNN architecture, and trigger geometry) from the given attack type. Taken collectively, these examples the flexibility of our approach: each attack is more easily identified by some metrics than others, and this varies between datasets and attack configurations. This highlights the importance of considering our metrics collectively rather than in isolation.

        \vspace{2mm}
        \subsubsection{Composite Metric}
        \label{sec:composite_metric}
        As no single metric is foolproof, using multiple metrics at once increases confidence in detection. Our composite metric evaluates to True if an individual graph meets the \textit{number of positive metrics required} (NPMR) to be classified as backdoored. Selecting an appropriate NPMR is critical for effective detection: setting NPMR too \textit{low} (e.g., 1) could lead to false detection, whereas setting it too \textit{high} (e.g., 7) could result in missed detection because of overly strict criteria. The intuition is that any given backdoor attack may not conform to the expectations of \emph{all} seven metrics, but it will likely conform to \emph{some}. (For example, an optimized trigger may not be a single connected graph, making the connectivity metric less useful.) The varied collection of detection metrics favors the likelihood that, regardless of attack type, at least some metrics will detect the backdoor, offering a robust and adaptable solution. While NPMR selection is critical, we have determined that several choices are good, finding that NPMR between 1 and 3 works well.

        \textbf{NPMR Selection Guidelines:} Our comprehensive evaluation across multiple datasets and attack types provides principled guidance for NPMR selection in practice. NPMR selection balances false positive and false negative rates based on application security requirements. We recommend starting with NPMR=2 as a robust default, which demonstrates consistent performance across diverse scenarios (average F1 scores of 0.707-0.786). For applications requiring higher sensitivity, NPMR=1 increases detection rates at the cost of potential false positives. When precision is prioritized over recall, NPMR=3-4 reduces false alarms while potentially missing some attacks. The specific choice should be validated against domain-specific clean validation data and adjusted based on the deployment threat environment.

    \subsection{Adaptive Backdoor Trigger}
        \label{sec:adaptive_attack}
    
        Since all metrics but NDV derive from the explanation process, a backdoor attack evading GNN explanation might also evade detection. To test our method's robustness, we propose an attack targeting both GNN classification and explanation. 
        The idea is to train a GNN to generator triggers evading explanation, while simultaneously training a separate, surrogate GNN to minimize classification loss on the poisoned graphs. Algorithm~\ref{alg:adaptive} presents this process; we detail these steps below.

        {\bf Note:} for the purposes of this algorithm, we momentarily revise the notation to represent edges with adjacency matrix $\textbf{A}$ rather than the set $\edges$. Conceptually, $\mask^A$ is the same as $\edgemask$, $\hat{\textbf{A}}$ is the adjacency matrix of a backdoored graph, and $\textbf{B}$ is the mask for edges not in the clean graph. 
        
        \begin{figure}[!t]
            \centering
            \includegraphics[width=\linewidth]{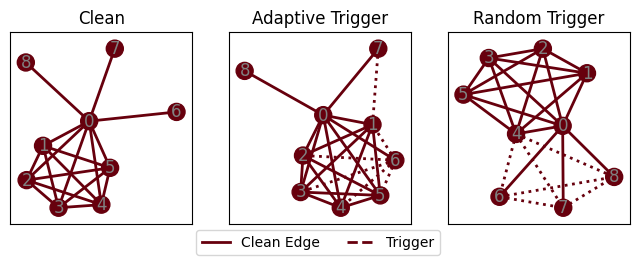}
            \caption{The graph with the adaptive trigger is much more faithful to the original structure than one with a pre-defined structure, such as a random trigger.}
            \label{fig:adaptive_trigger}
            \vspace{-4mm}
        \end{figure}

\newcommand{\fontchange}[1]{{\fontfamily{phv}\selectfont #1}}

\begin{algorithm}[t]
\footnotesize
    \SetAlgoLined
    \LinesNumbered
    \SetInd{0.5em}{0.5em}
    \caption{Training an Adaptive Trigger Generator}
    \label{alg:adaptive}
        \textbf{Input:} Untrained graph classifier \fso, untrained trigger edge generator \fgen, clean graph dataset $D$, a set of to-be-backdoored graphs $D_B \subset D$, with the target label $\hat{y}$, number of steps $T$, number of epochs \#Epochs, GNNExplainer learning rate $\eta_{exp}$, edge generator learning rate $\eta_{gen}$, trigger size $s$\\
        \textbf{Output:} Trained edge generator \fgen\\
        \fgen\ $\gets$ \fgeno; \\%
        \fs\ $\gets$ \textsc{Train}(\fso, \textit{D}); \\
        \While{not converged}{
            \For{\upshape{epoch}=1 to \#Epoch}{
            $L_{gen} \gets 0$; \\
                \For{$G=(\A,\X)$ \upshape{in} $D_B$}{
                    $\textbf{B} \gets \textbf{11}^\textrm{T}-\textbf{I}-$\A \tcp*{all non-existent edges}
                    $(u,v) \gets \text{argmax}_{u,v}$\fgen$($\A$,\X,\textbf{B})$; \tcp{non-existent edge w.r.t. maximum \fgen output}
                    $\Ahat = {\bf A}$, ${\bf \hat{A}}_{u,v}=1$ \tcp*{Add edge $(u,v)$ to ${\bf \hat{A}}$}
                    $\maskAhat_{0} \gets \text{Randomly initialize the mask on } \Ahat$;\\
                    \For{$t = 0$ \upshape{to} $T$}{
                        $L_{CE} \gets \textsc{CrossEntropyLoss}($\fs$(\Ahat \odot \maskAhat_{t}, \X), \hat{y})$; \\
                        $\maskAhat_{t+1} \gets \maskAhat_{t} - \eta_{exp} \frac{\partial{ L_{CE}}}{\partial \maskAhat_t}$; 
                        }                
                    $L_{gen} \gets L_{gen} + \sum_{i,j}(\maskAhat_{T} \odot \textbf{B})_{ij}$;
                    }
                $\theta_{gen} \gets \theta_{gen} - \eta_{gen} \frac{\partial L_{gen}}{\partial {\theta_{gen}}};$  \tcp{Update \fgen\ params}
                }
            $\hat{D}_B \gets \textsc{Poison}(D_B, f_{gen}, s)$ 
            \tcp*{Call \fgen\ for $s$ times on $D_B$ to generate backdoored graphs $\hat{D}_B$}
            $f_{s} \gets \textsc{Train}({f_s}^0, {D}\setminus D_B \cup \hat{D}_B)$ \tcp*{\scriptsize{Retrain \fs\ on updated data}}
            }
\end{algorithm}

            \begin{itemize}
                \item[1.] \textit{Begin with untrained edge generator GNN, and a surrogate graph classifier pre-trained on clean data.}

                Attackers begin with an untrained GNN trigger edge generator, \fgeno  (Algorthm~\ref{alg:adaptive} line 3), and an untrained GNN graph classifier, \fso. In the first stage (line 4), they train \fso\ on clean data to obtain \fs.

                \item[2.] \textit{Iteratively repeat the following process:}
                \begin{itemize}
                    \item[a.] \textit{Use edge generator to add a new edge to clean graphs.}

                    Let $D_B$ be a subset of clean graphs $D$ designated for attack. Each graph  $G=({\bf A}, {\bf X}) \in D_B$ is fed to \fgen\, which outputs a score for each (not-yet-existing) edge in \textbf{B}. 
                    These scores will correspond to the likelihood of an edge being \textit{excluded} by GNNExplainer — i.e., the stealthiness of each edge choice. Therefore, the edge with the highest score is the one marked for inclusion in the trigger (line 10).\\
                    
                    \item[b.] \textit{Obtain explanation of current classifier's prediction on modified graphs.}

                    Attackers obtain $\Ahat$, the adjacency matrix $\A$ updated with the added edge (line 11). They then simulate GNNExplainer's iterative process of optimizing $\maskAhat_t$: in each step $t$, they compute cross-entropy loss on \fs's prediction on a graph with edges weighted by $\maskAhat_t$; they then update $\maskAhat_t$ \emph{in the opposite direction} of the gradient with respect to $\maskAhat_t$ (line 15).
    
                    \item[c.] \textit{Perform gradient descent on generator.}

                    The objective function aims to find the \fgen\ whose edge predictions evade explanation by GNNExplainer:
                
                    \small
                        \begin{equation}
                            \resizebox{!}{8.5pt}{$
                            \min_{\hat{A}} \sum \maskAhat_t\odot \textbf{B}.
                            $}
                        \label{adaptive_loss}
                        \end{equation}
                    $\mask^{\hat{A}}\odot \textbf{B}$ represents the explanatory mask weights for the new edges. Minimizing this value trains the edge generator to produce stealthy edges in $\hat{\textbf{A}}$: those that the GNN explainer deems unimportant.
                    
                    For each graph, attackers update the running loss value, $L_{gen}$ with the summed explanation mask values associated with the newest edges (line 17). By masking $\maskAhat_t$ by \textbf{B} within this summation, attackers limit the influence on this loss term to non-clean edges only. After iterating over all to-be-backdoored graphs $D_B$, edge generator parameters $\theta_{gen}$ are updated in the opposite direction of the gradient (line 20).

                    \item[d.] \textit{Poison the dataset and retrain the surrogate graph classifier to minimize classification loss.}

                    For each graph in $D_B$ designated for backdoor, attackers call \fgen\ for $s$ times in order to inject a trigger with $s$ edges. Attackers then retrain the model \fs\ on this attacked dataset. This simulates the realistic scenario in which GNNExplainer is operating against an already-attacked GNN.
                \end{itemize}
            \end{itemize}
    The above process is repeated until $L_{gen}$ converges. In the end, attackers have an edge generator \fgen that can generate triggers of arbitrary size (one edge at a time). They then use this generator \fgen\ to generate adaptive triggers within a dataset. This process yields a stealthy trigger that blends in with clean graph geometry, as seen in Fig. \ref{fig:adaptive_trigger}.


\section{Experiments}
\label{sec:results}
We evaluate our method against a variety of backdoor attacks for multiple datasets and GNN architectures. 
    \subsection{Experimental Setup}

        \subsubsection{Datasets}
            Our six graph datasets were selected for their diverse structure and domain. MUTAG~\cite{debnath1991structure} contains chemical compounds labeled by their mutagenic effect.  AIDS~\cite{riesen2008iam} contains graphs of molecular structures tested for activity against HIV. PROTEINS~\cite{10.1093/bioinformatics/bti1007} includes enzyme and non-enzyme structures. Lastly, IMDB-BINARY and IMDB-MULTI~\cite{10.1145/2783258.2783417} are datasets of actors' ego-networks labeled by movie genre. IMDB-BINARY contains 1,000 networks from Action and Romance films for binary classification, while IMDB-MULTI (a distinct dataset) contains 1,500 networks from Comedy, Romance, and Sci-Fi films for multi-class classification. In both datasets, nodes are actors and edges connect co-starring actors.
            
            \begin{table}[!t]    
              \centering
              \footnotesize
              \begin{tabular}{@{}lcccc@{}}
                \toprule
                \multirow{2}{*}{Dataset} & \multicolumn{1}{p{1cm}}{\centering Num.\\ Classes} & \multicolumn{1}{p{1cm}}{\centering Graph\\ Count} & \multicolumn{1}{p{1cm}}{\centering Avg.\\ Nodes} & \multicolumn{1}{p{1cm}}{\centering Avg.\\ Edges} \\
                \midrule
                MUTAG        & 2 & 188   & 17.93 & 19.79 \\
                AIDS         & 2 & 2000  & 15.69 & 16.20 \\
                PROTEINS     & 2 & 1113  & 39.06 & 72.82 \\
                DBLP         & 2 & 5000  & 10.48 & 19.65 \\
                IMDB-BINARY  & 2 & 1000  & 19.77 & 96.53 \\
                IMDB-MULTI   & 3 & 1500  & 13.00 & 65.94 \\
                \bottomrule
              \end{tabular}
              \caption{Dataset Properties}   
              \label{tab:datasets}
              \vspace{-4mm}
            \end{table}

        \vspace{2mm}
        \subsubsection{Attack Models}
            We use a traditional random attack~\cite{zhang2021backdoor}, a clean-label random attack~\cite{10.1145/3548606.3563531}, Motif-Backdoor~\cite{10108961}, TRAP~\cite{Yang2022TransferableGB}, LIN~\cite{10.1145/3468218.3469046}, and our adaptive attack outlined above. (Note: while several attacks both utilize random methods, in this paper, ``random'' refers only to ~\cite{zhang2021backdoor}.

        \vspace{2mm}
        \subsubsection{Parameter Settings}
            \begin{itemize}
                \item \textbf{Trigger Sizes:} Random, clean-label, TRAP, LIN, and adaptive triggers ranged from 2 to 12 nodes.\footnote{Due to their large degrees, IMDB-BINARY and IMDB-MULTI required larger triggers (26–36 nodes), which were mapped to 2–12 for consistency.} Motif triggers, as per~\cite{10108961}, consist of 3 or 4 nodes (see Fig.~\ref{fig:motif_types}).
                \item \textbf{Inter-Connectivity:} Random, clean-label, and LIN triggers used inter-connectivity probabilities of 1\%, 50\%, or 100\%. TRAP trigger inter-connectivity varies implicitly as a function of available edge positions, gradient scores, and trigger size. Motif-based triggers had predefined inter-connectivity.
                \item \textbf{Poison Ratio:} Set at 20\% of training data to ensure sufficient poisoned samples for testing. Graphs for poisoning were sampled uniformly at random.
                \item \textbf{Classifier Architectures:} Used GCN~\cite{kipf2016semi}, GIN~\cite{xu2018how}, and GAT~\cite{velivckovic2017graph}, with 2–4 layers, 16–256 hidden dimensions, and training over 150–600 epochs.
                \item \textbf{GNNExplainer:} Set mask and edge entropy weights to 1, and mask and edge size weights to $1\text{e-}4$ to get decisive subgraphs that preserve backdoor features.
                \item \textbf{Trigger Generator:} Four fully connected layers (64 hidden dimensions). Trained for 20 epochs, optimizing explanatory masks over 50–100 steps per epoch. Learning rates for both generator and mask optimization ranged from 0.005 to 0.05.
            \end{itemize}
        \subsubsection{Hardware and Software Specifications} Experiments were run on a MacBook Pro with an Apple M1 Pro chip (8 cores: 6 performance, 2 efficiency) and 16 GB memory, on macOS Sonoma Version 14.5. Models were implemented in Python with the PyTorch framework.
       \vspace{-2mm}
        
    \subsection{Experimental Results} 
            Each detection attempt with a specific NPMR yields a confusion matrix, helping determine the NPMR that maximizes the detection F1 score. The goal is to maximize True Positives (correct detection) and minimize False Negatives (missed detection). The F1 score balances these, even when most samples are clean. 
            Table \ref{tab:comprehensive_results} shows the result. Each row represents the average F1 scores at different NPMRs across 544 random, 808 clean-label, 26 motif-based, 156 TRAP, 159 LIN, and 93 adaptive attacks.
            \footnote{We intended to explore GTA~\cite{xi2021graph}, but source code issues prevented it. However, detecting our adaptive attack--designed to challenge our method--demonstrates robustness against optimized attacks. Some popular methods are also irrelevant here; e.g., ``Unnoticeable'' Backdoor Attacks~\cite{10.1145/3543507.3583392} aren’t designed for graph-level classification, and PoisonedGNN~\cite{Alrahis2023ttPoisonedGNNBA} targets hardware security applications. Non-graph defenses, like STRIP~\cite{gao2020stripdefencetrojanattacks}, often don’t translate to graph data, as concepts like ``mixing'' samples lack clear graph equivalents.
            }

\begin{table}[h]
                \centering
                \resizebox{0.49\textwidth}{!}
                {
                \scriptsize
                \begin{tabular}{@{}lc|*{7}{c}|cc@{}}

                    \toprule

                    {} & {} & \multicolumn{7}{c|}{Our Method - F1 at NPMR} & \multicolumn{2}{c}{Baselines - Max F1} \\
                    & Dataset & 1 & 2 & 3 & 4 & 5 & 6 & 7 & ES & XGBD \\
                    \midrule
                    \multirow{7}{*}{\rotatebox[origin=c]{90}{Random}} 
                    & All & 0.717 & 0.786 & \textbf{0.809} & 0.727 & 0.493 & 0.187 & 0.029 & 0.537 & 0.608 \\
                    & MUTAG & 0.563 & 0.732 & 0.883 & \textbf{0.911} & 0.802 & 0.264 & 0.013 & 0.421 & 0.789 \\
                    & AIDS & 0.757 & \textbf{0.815} & 0.813 & 0.671 & 0.356 & 0.113 & 0.007 & 0.576 & 0.708 \\
                    & PROTEINS & 0.760 & 0.852 & \textbf{0.908} & 0.830 & 0.559 & 0.200 & 0.005 & 0.544 & 0.631 \\
                    & DBLP & 0.731 & \textbf{0.756} & 0.717 & 0.586 & 0.362 & 0.154 & 0.040 & 0.708 & 0.347 \\
                    & IMDB-BIN. & 0.695 & \textbf{0.704} & 0.648 & 0.557 & 0.402 & 0.199 & 0.031 & 0.435 & 0.566 \\
                    & IMDB-MULT. & 0.796 & 0.858 & \textbf{0.883} & 0.809 & 0.497 & 0.191 & 0.081 & 0.814 & 0.705 \\
                    \midrule
                    \multirow{7}{*}{\rotatebox[origin=c]{90}{Clean Label}} 
                    & All & \textbf{0.828} & 0.779 & 0.650 & 0.464 & 0.264 & 0.097 & 0.021 & 0.506 & 0.571 \\
                    & MUTAG & 0.628 & \textbf{0.717} & 0.714 & 0.639 & 0.410 & 0.133 & 0.017 & 0.447 & 0.680 \\
                    & AIDS & \textbf{0.883} & 0.777 & 0.571 & 0.331 & 0.104 & 0.022 & 0.002 & 0.554 & 0.520 \\
                    & PROTEINS & \textbf{0.862} & 0.747 & 0.560 & 0.331 & 0.148 & 0.027 & 0.002 & 0.530 & 0.696 \\
                    & DBLP & \textbf{0.884} & 0.719 & 0.482 & 0.273 & 0.125 & 0.033 & 0.006 & 0.635 & 0.446 \\
                    & IMDB-BIN. & \textbf{0.845} & 0.831 & 0.756 & 0.550 & 0.399 & 0.222 & 0.067 & 0.365 & 0.514 \\
                    & IMDB-MULT. & 0.865 & \textbf{0.884} & 0.814 & 0.662 & 0.400 & 0.143 & 0.031 & 0.606 & 0.570 \\
                    \midrule
                    \multirow{7}{*}{\rotatebox[origin=c]{90}{Motif}} 
                    & All & \textbf{0.744} & 0.707 & 0.558 & 0.331 & 0.149 & 0.053 & 0.012 & 0.286 & 0.193 \\
                    & MUTAG & \textbf{0.846} & 0.833 & 0.471 & 0.167 & 0.167 & 0.000 & 0.000 & 0.250 & 0.247 \\
                    & AIDS & 0.736 & \textbf{0.755} & 0.677 & 0.495 & 0.273 & 0.102 & 0.025 & 0.442 & 0.206 \\
                    & PROTEINS & 0.613 & \textbf{0.640} & 0.562 & 0.370 & 0.087 & 0.087 & 0.000 & 0.286 & 0.243 \\
                    & DBLP & \textbf{0.763} & 0.653 & 0.405 & 0.114 & 0.006 & 0.000 & 0.000 & 0.165 & 0.074 \\     
                    & IMDB-BIN. & n/a & n/a & n/a & n/a & n/a & n/a & n/a & n/a & n/a \\
                    & IMDB-MULT. & n/a & n/a & n/a & n/a & n/a & n/a & n/a & n/a & n/a \\
                    \midrule
                    \multirow{7}{*}{\rotatebox[origin=c]{90}{TRAP}} 
                    & 
                    All & 
                    0.783 & \textbf{0.790} & 0.714 & 
                    0.578 & 0.366 & 0.183 & 
                    0.076 & 
                    0.617 & 0.450 \\
                    & 
                    MUTAG & 
                    0.679 & 0.692 & 0.725 & 
                    \textbf{0.786} & 0.252 & 0.100 & 
                    0.000 & 
                    0.031 & 0.426 \\
                    & 
                    AIDS &
                    \textbf{0.838} & 0.834 &
                    0.750 & 0.587 &
                    0.377 & 0.132 &
                    0.001 & 
                    0.562 & 0.458 \\
                    & 
                    PROTEINS &
                    0.751 & \textbf{0.796} &
                    0.754 & 0.636 &
                    0.440 & 0.202 &
                    0.029 & 
                    0.453 & 0.531 \\
                    & 
                    DBLP &
                    \textbf{0.872} & 
                    0.842 &
                    0.743 & 
                    0.531 &
                    0.352 & 
                    0.162 &
                    0.038 & 
                    0.544 & 
                    0.411 \\
                    & 
                    IMDB-BINARY &
                    0.711 & 
                    0.627 &
                    0.518 & 
                    0.409 &
                    0.299 & 
                    0.175 &
                    0.114 & 
                    \textbf{0.786} & 
                    0.337 \\
                    & 
                    IMDB-MULTI &
                    0.755 & 
                    0.796 &
                    0.688 & 
                    0.536 &
                    0.379 & 
                    0.269 &
                    0.183 & 
                    \textbf{0.875} & 
                    0.536 \\
                    \midrule
                    \multirow{7}{*}{\rotatebox[origin=c]{90}{LIN}} 
                    &
                    All &
                    0.750 & \textbf{0.797} &
                    0.787 & 0.660 &
                    0.423 & 0.190 &
                    0.060 & 
                    0.569 & 
                    {0.741} \\
                    &
                    MUTAG &
                    0.707 &
                    0.758 &
                    \textbf{0.817} &
                    0.764 &
                    0.220 &
                    0.006 &
                    0.000 & 
                    0.335 & 0.733 \\
                    &
                    AIDS &
                    0.778 & \textbf{0.826} &
                    0.789 & 0.623 &
                    0.373 & 0.147 &
                    0.031 & 
                    0.666 & 0.811 \\
                    &
                    PROTEINS &
                    0.740 & 0.796 &
                    \textbf{0.798} & 0.664 &
                    0.432 & 0.142 &
                    0.016 & 
                    0.466 & 0.706 \\
                    &
                    DBLP &
                    0.753 & \textbf{0.784} &
                    0.745 & 0.605 &
                    0.407 & 0.214 &
                    0.070 & 
                    0.641 & 0.661 \\
                    &
                    IMDB-BINARY &
                    0.543 & 0.654 &
                    0.734 & \textbf{0.764} &
                    0.636 & 0.484 &
                    0.297 & 
                    0.438 & 0.604 \\
                    &
                    IMDB-MULTI &
                    0.894 & 0.894 &
                    \textbf{0.897} & 0.897 &
                    0.863 & 0.789 &
                    0.565 & 
                    0.867 & 0.867 \\
                    \midrule
                    \multirow{7}{*}{\rotatebox[origin=c]{90}{Adaptive}}
                    & All & 0.695 & \textbf{0.723} & 0.699 & 0.616 & 0.380 & 0.179 & 0.041 & 0.569 & 0.392 \\
                    & MUTAG & 0.551 & 0.689 & 0.820 & \textbf{0.859} & 0.557 & 0.186 & 0.000 & 0.680 & 0.481 \\
                    & AIDS & 0.755 & \textbf{0.815} & 0.801 & 0.659 & 0.307 & 0.041 & 0.006 & 0.693 & 0.348 \\
                    & PROTEINS & 0.733 & \textbf{0.744} & 0.704 & 0.601 & 0.295 & 0.061 & 0.004 & 0.503 & 0.195 \\
                    & DBLP & \textbf{0.697} & 0.657 & 0.590 & 0.440 & 0.184 & 0.092 & 0.016 & 0.492 & 0.689 \\
                    & IMDB-BIN. & \textbf{0.680} & 0.618 & 0.490 & 0.361 & 0.278 & 0.193 & 0.106 & 0.475 & 0.247 \\
                    & IMDB-MULT. & 0.753 & \textbf{0.804} & 0.786 & 0.775 & 0.658 & 0.504 & 0.115 & 0.571 & 0.535 \\
                    \bottomrule
                \end{tabular}
                }
                \caption{Detection F1 scores at varied NPMR and comparison with baseline methods. Bold values indicate the best performance for our method on each dataset. Our method consistently outperforms both ES~\cite{Jiang2022DefendingAB} and XGBD~\cite{guan2023xgbd} baselines across all attack types and datasets.}
                \label{tab:comprehensive_results}
                \vspace{-6mm}
            \end{table}

        \vspace{2mm}
        \subsubsection{Detection F1 Scores for Varied Attacks}
            
             We test our detection method against random, clean-label, motif-based, TRAP, LIN, and adaptive attacks.

            \begin{itemize}[leftmargin=*]
                \item{\textbf{Random Attack.} Using the metric of the highest-attained F1 score, our method is most successful against the traditional random attack (best-case F1=0.911). This is unsurprising: since trigger generation and insertion are randomized, there is no attempt by the attacker to be stealthy. Average F1 scores peak at NPMR=3, indicating that triggers generated by random attacks follow our assumptions for backdoor behavior along multiple metrics.}
                \item{\textbf{LIN.} The highest F1 score against LIN attacks is 0.897, making it the second-best detected attack after random attacks. While this attack aims to be stealthy by strategically targeting the least-important nodes for trigger placement, the random trigger geometry still creates detectable signatures that our metrics can identify. The relatively high detection performance compared to our adaptive attack demonstrates that optimizing placement alone is insufficient when trigger patterns retain their artificial, random characteristics.}
                \item{\textbf{Clean-Label Attack.} The highest F1 score our detection method attains against a clean-label attack is 0.884, which is slightly lower than performance on the traditional random attack. Clean-label attacks are more covert than traditional random attacks of the same trigger size, as the poisoned graphs closely resemble the target class. This is likely why the average F1 score peaks earlier, at NPMR=1: the poisoned graphs differ from clean ones in fewer aspects compared to traditional random attacks.}
                \item{\textbf{TRAP.} The highest F1 score against TRAP attacks is 0.872, ranking 4th among all tested attacks. TRAP's perturbation-based approach, which generates sample-specific triggers via gradient score matrices, presents moderate detection challenges. Unlike discrete subgraph injection methods, TRAP's perturbations create more subtle signatures that still remain detectable through our explanation-based metrics.}
                \item{\textbf{Adaptive Attack.} The bes  F1 score against our adaptive attack is 0.859, lower than the random and clean-label results, as expected, since the adaptive attack is specifically designed to challenge our detection method by generating triggers that carefully evade the metrics we use. Despite this, our detection remains robust, as evidenced by reasonably high F1 scores and an average F1 peak at NPMR=2. These results underscore the strength of our adaptive attack and the resilience of our method against such tailored threats.}
                \item{\textbf{Motif-Backdoor.} The best-case F1 score against Motif-Backdoor is 0.846, with average F1 peaking at NPMR=1. While this detection score is lower than that for the adaptive attack, it does not undermine our assertion that the adaptive attack poses the strongest threat to our method. The lower score for the motif attack is attributable to its unique design, which uses small triggers (3 to 4 nodes) that are subtler and inherently harder to detect compared to the larger triggers (up to 12 nodes) employed by the other tested attacks. These small triggers also limit the motif attack’s ability to meaningfully influence predictions in our datasets. 
                As a result, IMDB-BINARY and IMDB-MULTI are excluded from motif-backdoor results due to the attack’s insufficient effectiveness on these datasets' dense graphs.}
            \end{itemize}
            Note that using NPMR=2 generalizes well across all attacks, producing average F1 ranging from 0.707 to 0.797.\\

            \subsubsection{Comparison against Baseline Detection Methods}
            Table~\ref{tab:comprehensive_results} also compares our method's effectiveness against XGBD \cite{guan2023xgbd} and Jiang et al.'s explanation-based method \cite{Jiang2022DefendingAB}, ES, our detection baselines.
            Note that ES is one of our seven metrics, and therefore, our method should be at least as successful as that of Jiang et al. Moreover, like Guan et al., our method hypothesizes distinct behavior between explanation losses on clean and backdoor samples; since our method not only shares this insight, but also incorporates other insights, we anticipate that our method will meet or exceed the effectiveness of XGBD. 
            The results presented in Table~\ref{tab:comprehensive_results} largely confirm these expectations. Across 36 attack type/dataset combinations tested, our method outperforms both Jiang et al. (ES) and Guan et al. (XGBD) in 34 cases. The two exceptions occur with TRAP attacks on IMDB-BINARY and -MULTI, where ES achieves higher F1 scores (0.786 and 0.875 respectively) than our composite approach (0.711 and 0.755), perhaps because TRAP's gradient-based perturbations create particularly strong explainability signatures in dense social networks that ES captures effectively as a standalone metric. For all other attack types across all datasets, and for TRAP on the remaining four datasets, our composite metric approach shows superior detection performance.

        \vspace{2mm}            
        \subsubsection{Detection Accuracy for Varied Attacks}
        {To provide a more holistic view of our method's effectiveness, Table~\ref{table:accuracy} presents detection performance in terms of accuracy. The values in bold correspond to the NPMR where average F1 is maximized.}
        
        \begin{table}[!t]
            \centering
            \resizebox{0.49\textwidth}{!}
            {
            \begin{tabular}{l|llllllll}
                \toprule
                    & {} & {} & {} & NPMR & {} & {} \\
                    & 1 & 2 & 3 & 4 & 5 & 6 & 7 \\
                \midrule
                    Random & 0.667 & 0.801 & \textbf{0.814} & 0.748 & 0.609 & 0.541 & 0.532 \\
                    Clean Label & \textbf{0.762} & 0.566 & 0.390 & 0.261 & 0.177 & 0.136 & 0.129 \\
                    Motif & \textbf{0.675} & 0.703 & 0.595 & 0.536 & 0.481 & 0.445 & 0.422 \\
                    TRAP & 0.689 & \textbf{0.731} & 0.693 & 0.627 & 0.533 & 0.470 & 0.436\\
                    LIN & 0.673 & \textbf{0.771} & 0.795 & 0.729 & 0.625 & 0.550 & 0.516\\
                    Adaptive & 0.640 & \textbf{0.756} & 0.761 & 0.717 & 0.605 & 0.561 & 0.550 \\
                \bottomrule
            \end{tabular}
            }
            \caption{Average detection accuracy for Random, Clean Label, Motif, TRAP, LIN, and Adaptive attacks at varied NPMR thresholds.}
            \label{table:accuracy}
            \vspace{-6mm}
        \end{table}

        These results match our earlier analysis of F1 scores: detection accuracy is highest for random attacks, and lowest for motif-based attacks.\footnote{The only difference is that the best-case detection accuracy for TRAP is lower than the same measurement for the adaptive attack, unlike their corresponding F1 scores, where TRAP was higher. One theory is that TRAP's gradient-optimized patterns occasionally overlap with clean graph variations, creating scenarios where there may be false backdoor alarms on clean inputs. This would make F1 a more flattering detection metric for TRAP since it handles false positives better than detection accuracy.} For all metrics, increasing NPMR eventually reduces accuracy and F1. This is likely because higher NPMR introduces stricter criteria for classification, thus increasing the rates of missed backdoor detections.

    \vspace{2mm}
    \subsubsection{Performance using Other Explainer Algorithms}
    We previously established in Fig. \ref{fig:gnn_inconsistency} that GNNExplainer, PGExplainer, and CaptumExplainer all fail to consistently and sufficiently reveal backdoor triggers in their explanatory subgraphs. Moreover, PGExplainer and CaptumExplainer do not support all 7 of our detection metrics: PGExplainer lacks node masks needed for Connectivity and SNDV, and CaptumExplainer lacks gradient-based optimization needed for loss curve metrics (Curvature and Elbow). 

        \begin{table}[h]
            \begin{tabular}{@{}lc|*{8}{c}@{}}
                \toprule
                {} & {} & \multicolumn{7}{c}{NPMR}\\
                & Dataset & 1 & 2 & 3 & 4 & 5 & 6 & 7\\
                \midrule
                \multirow{5}{*}{\rotatebox[origin=c]{90}{PGExpl.}} 
                & MUTAG    & 0.62 & 0.76 & 0.75 & 0.66 & 0.35 & n/a & n/a \\
                & AIDS     & 0.70 & 0.73 & 0.67 & 0.44 & 0.05 & n/a & n/a \\
                & PROTEINS & 0.64 & 0.68 & 0.63 & 0.27 & 0.03 & n/a & n/a \\
                & DBLP     & 0.67 & 0.66 & 0.41 & 0.13 & 0.01 & n/a & n/a \\
                & IMDB-BIN. & 0.60 & 0.61 & 0.46 & 0.13 & 0.00 & n/a & n/a \\
                & IMDB-MULT. & 0.67 & 0.77 & 0.83 & 0.52 & 0.08 & n/a & n/a \\
                \midrule
                \multirow{5}{*}{\rotatebox[origin=c]{90}{CaptumExpl.}} 
                & MUTAG    & 0.59 & 0.74 & 0.77 & 0.65 & 0.38 & n/a & n/a \\
                & AIDS     & 0.70 & 0.75 & 0.69 & 0.39 & 0.12 & n/a & n/a \\
                & PROTEINS & 0.67 & 0.71 & 0.65 & 0.48 & 0.09 & n/a & n/a \\
                & DBLP     & 0.67 & 0.65 & 0.50 & 0.18 & 0.03 & n/a & n/a \\
                & IMDB-BIN. & 0.59 & 0.54 & 0.29 & 0.06 & 0.00 & n/a & n/a \\
                & IMDB-MULT. & 0.70 & 0.73 & 0.57 & 0.27 & 0.04 & n/a & n/a \\          
                \bottomrule
            \end{tabular}
        \caption{PGExplainer and CaptumExplainer detection effectiveness (F1 scores) against random attacks.}
        \label{table:pg_captum_performance}
        \end{table}
    
    Since PGExplainer and CaptumExplainer support fewer detection metrics, we expect them to be less effective at detecting backdoor samples using our method. To test this theory, we tried using PGExplainer and CaptumExplainer to detect random backdoor attacks. Table~\ref{table:pg_captum_performance} shows the resulting F1 detection scores at varied NPMR. As expected, the fewer available metrics weaken detection effectiveness: the maximum F1 scores achieved using PGExplainer and CaptumExplainer are 0.83 and 0.77, respectively, versus the F1 score of 0.911 achieved by GNNExplainer. Even with access to only 5 of our 7 metrics, these alternate explainers still achieve respectable performance, suggesting our core insights are broadly applicable. However, the drop in performance reinforces our decision to build around GNNExplainer, whose richer outputs enable the full expressiveness of our detection framework.

    \subsection{Ablation Studies} 

        \subsubsection{Attack Strength vs. Composite F1 Score}

\begin{figure}[!t]
            \centering
            \includegraphics[width=0.9\linewidth]{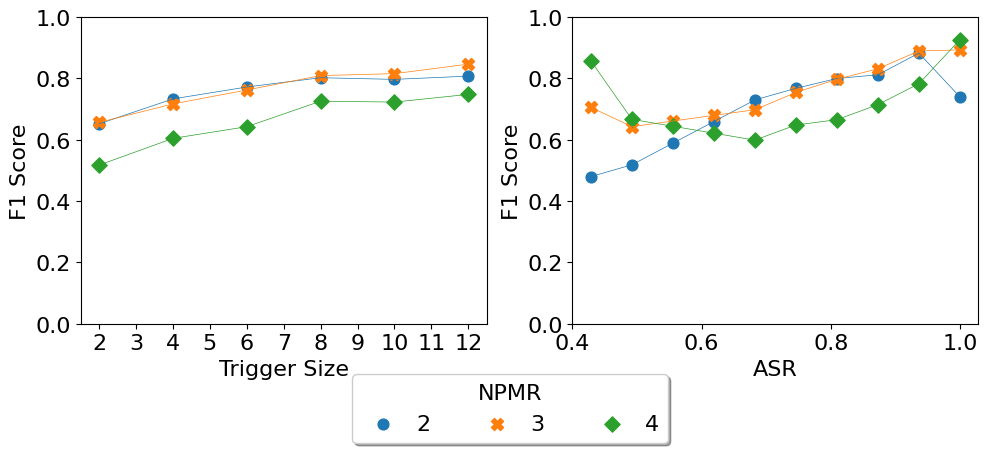}
            \caption{Effect of trigger size and attack success rate (ASR) of random attacks on composite metric F1 scores.}
            \label{fig:trigger_asr}
            \vspace{-6mm}
        \end{figure}

        Fig. \ref{fig:trigger_asr} plots detection F1 scores against trigger size and ASR for random, clean-label, and adaptive attacks, showing how these trends vary for NPMR ranging between 2 and 4. (Note: we exclude Motif Backdoor results from this figure, since motif sizes range only from 3 to 4.) These results suggest that stronger attacks may be easier to detect. Additionally, the figure demonstrates that lower NPMR values correlate with higher F1 scores. This is intuitive, as F1 places significant emphasis on true positives, and a higher NPMR reduces the likelihood of true positive detection by imposing stricter detection criteria.

        \vspace{2mm}
        \subsubsection{Effect of Poison Rate on ASR and Detection Performance}
        
        To evaluate whether our 20\% poisoning rate might artificially inflate detection performance, we evaluated our method using a 10\% poisoning rate and compared both attack success rates (ASR) and detection effectiveness.

        \begin{table}[htbp]
            \centering
            \resizebox{0.49\textwidth}{!}{
                \begin{tabular}{l|lllllll|l}
                \toprule
                    \multirow{2}{*}{Attack Type} & \multicolumn{7}{c|}{F1 at varied NPMR} & \multirow{2}{*}{ASR} \\
                     & 1 & 2 & 3 & 4 & 5 & 6 & 7 & \\
                    \midrule
                    Random 10\% & 0.70 & 0.76 & 0.77 & 0.71 & 0.50 & 0.21 & 0.03 & 0.63 \\
                    Random 20\% & 0.73 & 0.81 & 0.79 & 0.66 & 0.31 & 0.05 & 0.00 & 0.66 \\
                    \hline
                    Clean-Label 10\% & 0.82 & 0.73 & 0.58 & 0.40 & 0.23 & 0.07 & 0.01 & 0.57 \\
                    Clean-Label 20\% & 0.86 & 0.69 & 0.48 & 0.28 & 0.11 & 0.02 & 0.00 & 0.55 \\
                    \hline
                    Motif 10\% & 0.71 & 0.75 & 0.71 & 0.56 & 0.30 & 0.11 & 0.01 & 0.73 \\
                    Motif 20\% & 0.78 & 0.76 & 0.57 & 0.39 & 0.19 & 0.08 & 0.00 & 0.71 \\
                    \hline
                    TRAP 10\% & 0.75 & 0.77 & 0.66 & 0.51 & 0.34 & 0.18 & 0.08 & 0.75\\
                    TRAP 20\% & 0.77 & 0.79 & 0.70 & 0.56 & 0.36 & 0.19 & 0.08 & 0.77\\
                    \hline
                    LIN 10\% & 0.70 & 0.76 & 0.75 & 0.64 & 0.43 & 0.17 & 0.03 & 0.80\\
                    LIN 20\% & 0.75 & 0.80 & 0.79 & 0.66 & 0.42 & 0.19 & 0.06 & 0.82\\
                    \hline                    
                    Adaptive 10\% & 0.70 & 0.74 & 0.70 & 0.59 & 0.40 & 0.21 & 0.08 & 0.67 \\
                    Adaptive 20\% & 0.70 & 0.75 & 0.70 & 0.59 & 0.28 & 0.08 & 0.01 & 0.61 \\
                    \bottomrule
                \end{tabular}
            }
        \caption{Effect of Poisoning Rate on Attack Success and Detection Performance}
        \label{tab:poisoning_rate_comparison}
        \end{table}
        
        Table~\ref{tab:poisoning_rate_comparison} reveals several important findings: First, attack success rates remain relatively stable or even slightly increase for some attack types (Clean-Label, Motif, Adaptive), demonstrating that backdoor attacks maintain effectiveness even with fewer poisoned samples. This validates that 10\% poisoning rates represent realistic threat scenarios.
        
        Our detection method shows resilience to the reduced poisoning rate, and the overall detection capability remains strong. Random attacks maintain F1 scores up to 0.77 on average, and even our most challenging adaptive attacks achieve F1 scores up to 0.74. These results demonstrate that our detection method remains viable at realistic attack scenarios with lower poisoning rates.

        \vspace{2mm}
        \subsubsection{Prevalence of Individual Metrics}
\begin{figure}[H]
                \centering
                \vspace{-4mm}
                \includegraphics[width=0.9\linewidth]{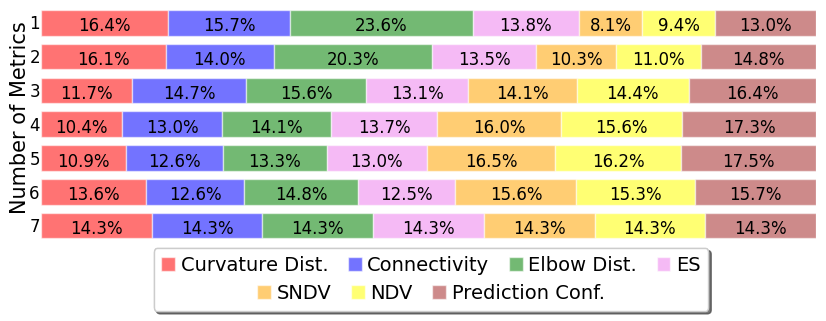}
                \vspace{-2mm}
                \caption{Inclusion rate of each metric among exactly $k$ positives, measured across all datasets and attack types.}
                \label{fig:saves}
                \vspace{-2mm}
            \end{figure}

            Fig. \ref{fig:saves} shows the percentage of detection attempts where each metric is among $k$ positive metrics, quantifying its influence in backdoor detection for different NPMRs. For example, SNDV is among exactly 1 positive metrics only 8.1\% of the time, but among 4 positive metrics 15.6\% of the time.
            This suggests the contributions of each metric are not consistent across NPMR.

            To better understand dataset and attack-specific patterns underlying these variations, we analyze which metrics most frequently contribute to detection when exactly two metrics are positive (the boundary cases for our recommended NPMR=2 threshold). Table~\ref{tab:metric_contribution} reveals important insights about metric complementarity across different scenarios:

            \begin{table}[h]
                \centering
                \vspace{-4mm}
                \setlength{\fboxrule}{2pt}
                    \begin{minipage}[b]{0.49\textwidth}
                        \centering
                        \begin{minipage}[b]{\textwidth}
                            \centering
                            \vspace{0.1cm}
                            \resizebox{\textwidth}{!}{
                                \begin{tabular}{@{}l|*{7}{@{\hspace{3pt}}c@{\hspace{3pt}}}cc@{}}
                            \toprule
                            Dataset & \rotatebox{45}{Curvature} & \rotatebox{45}{Connectivity} & \rotatebox{45}{Elbow} & \rotatebox{45}{ES} & \rotatebox{45}{SNDV} & \rotatebox{45}{NDV} & \rotatebox{45}{Pred. Conf.} \\
                            \midrule
                            MUTAG & \cellcolor{orange!45}13.1\% & \cellcolor{orange!60}17.2\% & \cellcolor{orange!60}17.4\% & \cellcolor{orange!45}14.8\% & \cellcolor{orange!45}12.9\% & \cellcolor{orange!45}12.2\% & \cellcolor{orange!45}12.4\% \\
                            AIDS & \cellcolor{orange!45}14.9\% & \cellcolor{orange!45}14.7\% & \cellcolor{orange!60}17.6\% & \cellcolor{orange!45}12.6\% & \cellcolor{orange!45}13.4\% & \cellcolor{orange!45}14.8\% & \cellcolor{orange!45}12.0\% \\
                            PROTEINS & \cellcolor{orange!45}14.2\% & \cellcolor{orange!75}19.4\% & \cellcolor{orange!60}17.6\% & \cellcolor{orange!60}17.1\% & \cellcolor{orange!30}10.1\% & \cellcolor{orange!30}10.3\% & \cellcolor{orange!30}11.3\% \\
                            DBLP & \cellcolor{orange!75}19.2\% & \cellcolor{orange!30}10.7\% & \cellcolor{orange!90}23.9\% & \cellcolor{orange!45}12.6\% & \cellcolor{orange!15}9.0\% & \cellcolor{orange!15}8.7\% & \cellcolor{orange!60}16.0\% \\
                            IMDB-BIN & \cellcolor{orange!30}10.6\% & \cellcolor{orange!60}15.7\% & \cellcolor{orange!60}17.8\% & \cellcolor{orange!60}15.0\% & \cellcolor{orange!30}10.8\% & \cellcolor{orange!45}13.7\% & \cellcolor{orange!60}16.4\% \\
                            IMDB-MULT & \cellcolor{orange!15}6.7\% & \cellcolor{orange!75}19.1\% & \cellcolor{orange!15}9.4\% & \cellcolor{orange!45}14.6\% & \cellcolor{orange!15}9.7\% & \cellcolor{orange!45}14.2\% & \cellcolor{orange!90}26.3\% \\
                            \bottomrule
                                \end{tabular}
                            }
                            
                            \vspace{2mm}
                            \small{(a) Importance by dataset, aggregated across attacks}
                        \end{minipage}
                        
                        \vspace{2mm} 
                        
                        \begin{minipage}[b]{\textwidth}
                            \centering
                            \vspace{0.1cm}
                            \resizebox{\textwidth}{!}{
                                \begin{tabular}{@{}l|*{7}{@{\hspace{3pt}}c@{\hspace{3pt}}}cc@{}}
                            \toprule
                            Attack Type & \rotatebox{45}{Curvature} & \rotatebox{45}{Connectivity} & \rotatebox{45}{Elbow} & \rotatebox{45}{ES} & \rotatebox{45}{SNDV} & \rotatebox{45}{NDV} & \rotatebox{45}{Pred. Conf.} \\
                            \midrule
                            Random & \cellcolor{orange!75}18.6\% & \cellcolor{orange!60}17.4\% & \cellcolor{orange!75}20.7\% & \cellcolor{orange!60}15.0\% & \cellcolor{orange!15}9.2\% & \cellcolor{orange!15}9.9\% & \cellcolor{orange!15}9.2\% \\
                            Clean Label & \cellcolor{orange!60}15.8\% & \cellcolor{orange!45}12.9\% & \cellcolor{orange!75}21.1\% & \cellcolor{orange!45}12.8\% & \cellcolor{orange!30}10.4\% & \cellcolor{orange!30}10.9\% & \cellcolor{orange!60}16.1\% \\
                            Motif & \cellcolor{orange!45}12.6\% & \cellcolor{orange!60}16.5\% & \cellcolor{orange!60}17.5\% & \cellcolor{orange!45}12.9\% & \cellcolor{orange!30}11.3\% & \cellcolor{orange!45}13.3\% & \cellcolor{orange!60}15.9\% \\
                            Adaptive & \cellcolor{orange!90}22.1\% & \cellcolor{orange!60}15.6\% & \cellcolor{orange!90}22.1\% & \cellcolor{orange!30}11.7\% & \cellcolor{orange!15}8.3\% & \cellcolor{orange!15}9.3\% & \cellcolor{orange!30}11.0\% \\
                            TRAP & \cellcolor{orange!45}14.4\% & \cellcolor{orange!45}13.4\% & \cellcolor{orange!60}16.0\% & \cellcolor{orange!60}17.9\% & \cellcolor{orange!30}11.5\% & \cellcolor{orange!30}10.6\% & \cellcolor{orange!60}16.2\% \\
                            LIN & \cellcolor{orange!60}16.5\% & \cellcolor{orange!60}16.6\% & \cellcolor{orange!60}17.6\% & \cellcolor{orange!45}14.7\% & \cellcolor{orange!30}10.7\% & \cellcolor{orange!30}11.6\% & \cellcolor{orange!45}12.3\% \\
                            \bottomrule
                                \end{tabular}
                            }
                            \\[2mm]
                            \small{(b) Importance by attack, aggregated across datasets}  
                        \end{minipage}
                    \end{minipage}%
                \caption{Metric contribution analysis for backdoor detection. Heat map shows frequency (\%) of each metric appearing among the two most effective metrics when exactly two metrics exceed detection thresholds, representing boundary cases for our recommended NPMR=2 setting and demonstrating complementary effectiveness patterns across datasets and attack types.}
                \label{tab:metric_contribution}
                \vspace{-2mm}
            \end{table}

            Random and adaptive attacks rely most on Elbow (20.7\%, 22.1\%) and Curvature (18.6\%, 22.1\%). Random attacks create simple optimization targets while adaptive attacks still require detectable trigger strength for effectiveness.
            
            Clean label attacks show higher prediction confidence importance since their triggers maintain semantic consistency: clean-label triggers inject into target classes, making models more confident than with semantically meaningless random or adaptive triggers.
            
            TRAP attacks are best detected by ES (17.9\%). This pattern suggests TRAP's gradient-optimized perturbations produce detectable changes in explanation fidelity.            
            
            LIN and motif-based attacks are best detected by Elbow (17.5\% and 17.6\%). This suggests that even for triggers that are strategically placed at low-importance locations, or that are designed to resemble naturally-occurring structures, the optimization process still creates detectable patterns in explanation loss curves.
            
            All attacks show similar connectivity importance, likely because many triggers use connected structures but explanations often capture mixed clean/backdoor components, diluting connectivity signatures.
            
            Dense social networks (IMDB) prioritize connectivity-based detection since high edge density makes geometric changes more detectable. Sparse citation networks (DBLP) focus on loss curve metrics since sparse structures are less affected by geometric changes. Molecular datasets show diverse approaches reflecting varied structural properties.

            \vspace{2mm}
            \subsubsection{Effect of Clean Validation Thresholding on Performance} 
            \label{sec:thresholding_effects}
            
            As mentioned in Section \ref{sec:detection_strategy}, we used the 25th/75th percentile of clean validation data to establish the thresholds for making a positive prediction for individual metrics.

            Table \ref{tab:percentiles} shows how the optimal NPMR changes depending on the thresholding settings. “Percentile” here refers to the \textit{upper} percentile — for example, a “95” threshold means a lower/upper threshold at the 5th/95th percentile of clean validation values. At the 50th percentile, a metric only has to beat the median of clean validation values to predict backdoor, which isn’t reasonable, because that includes 50\% of clean samples. For these lenient thresholds, higher NPMRs are appropriate, requiring several positive metrics to ensure confidence and minimize false alarms.

            \begin{table}[H]
                \centering
                \begin{tabular}{@{}c|c*{12}{c}@{}}
                    \toprule
                        {} & & & \multicolumn{7}{c}{Percentile} \\
                        Attack Type
                        & \multicolumn{1}{p{0.1cm}}{\centering 50} 
                        & \multicolumn{1}{p{0.1cm}}{\centering 55} 
                        & \multicolumn{1}{p{0.1cm}}{\centering 60} 
                        & \multicolumn{1}{p{0.1cm}}{\centering 65} 
                        & \multicolumn{1}{p{0.1cm}}{\centering 70} 
                        & \multicolumn{1}{p{0.1cm}}{\centering 75} 
                        & \multicolumn{1}{p{0.1cm}}{\centering 80} 
                        & \multicolumn{1}{p{0.1cm}}{\centering 85} 
                        & \multicolumn{1}{p{0.1cm}}{\centering 90} 
                        & \multicolumn{1}{p{0.1cm}}{\centering 95} 
                        & \multicolumn{1}{p{0.1cm}}{\centering 100}\\ 
                        \hline
                        Random          & 4 & 4 & 4 & 3 & 3 & \textbf{3} & 2 & 1 & 1 & 1 & 1 \\
                        Clean-Label     & 1 & 1 & 1 & 1 & 1 & \textbf{1} & 1 & 1 & 1 & 1 & 1 \\
                        Motif           & 4 & 4 & 3 & 3 & 3 & \textbf{2} & 2 & 1 & 1 & 1 & 1 \\
                        TRAP   & 3 & 3 & 2 & 2 &
                        2 & \textbf{1} & 1 & 1 & 1 & 1 & 1 \\
                        LIN & 4 & 4	& 3 & 3	& 
                        3 & \textbf{2}	& 2 & 1 & 1 & 1 & 1 \\
                        Adaptive            & 4 & 4 & 3 & 3 & 3 & \textbf{2} & 2 & 1 & 1 & 1 & 1\\

                        \bottomrule
                \end{tabular}
                \caption{Optimal NPMRs at varying clean validation thresholds, as determined by the maximum average F1 score. For our default 25th/75th percentile threshold (in bold), this value ranges from 1 to 3.} 
                \label{tab:percentiles}
            \end{table}
            
            When the threshold becomes stricter (e.g., 95th percentile), exceeding it provides stronger evidence of a backdoor trigger. In such cases, all attack types eventually favor NPMR=1, as a single metric surpassing this high threshold is sufficiently convincing to indicate the presence of a backdoor trigger. Note that for clean-label attacks, the optimal NPMR is 1 at every threshold. This is likely because clean-label triggers are subtle — backdoor patterns are embedded within the target class without changing the label — making a single positive metric sufficient to detect the differences, while requiring more metrics may result in false backdoor predictions. On the other hand, because random, TRAP, LIN, adaptive, and motif-based attacks inject triggers into both classes, they yield more varied differences between clean and backdoor samples; as a result, they rely on higher NPMR to identify backdoor samples under weaker thresholds.

        \subsection{Computational Complexity}
        Our method consists of \emph{1) running a GNN explainer} and \emph{2) computing a composite of 7 metrics using results of the explanation process}. The computational complexity of 2) is computed in constant time, and is negligible compared with 1). Therefore, the computational complexity of our method is determined by the explainer algorithm selected.


\section{Conclusion}
\label{sec:conclusion}
We explore the vulnerabilities of Graph Neural Networks (GNNs) to backdoor attacks and the challenges in detecting these intrusions, particularly given the general shortage of effective detection methods. Our research highlights the limits of using traditional GNN explainer outputs (i.e., explanatory subgraphs) in consistently revealing the full scope of backdoor information. To address these issues, we proposed a novel detection strategy that leverages seven new metrics, offering a more robust approach to backdoor detection. We demonstrate our method's effectiveness through extensive evaluations on various datasets and attack models. This robustness positions our approach as a valuable tool for strengthening GNN defenses against backdoor attacks.




\appendix

 \begin{figure*}[!t]
            \includegraphics[width=1\linewidth,trim=0 0 0 0, clip]{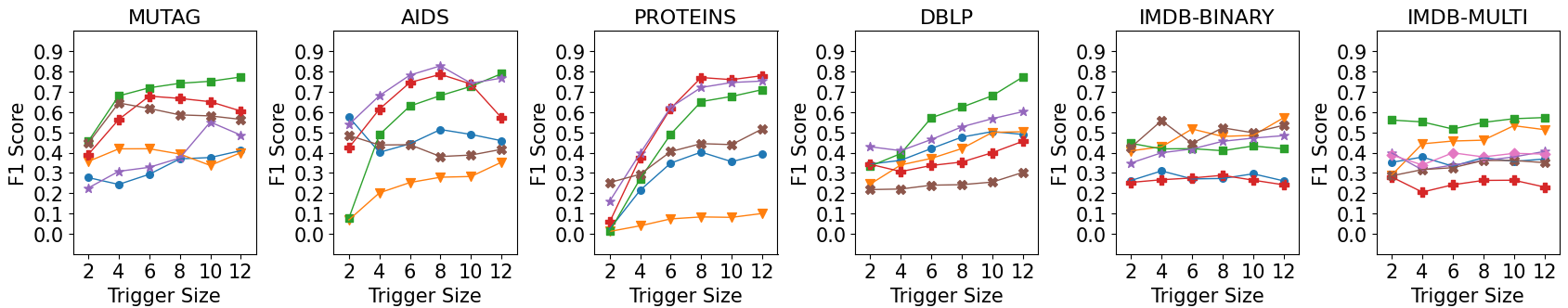}
            \includegraphics[width=1\linewidth,trim=0 0 0 0, clip]{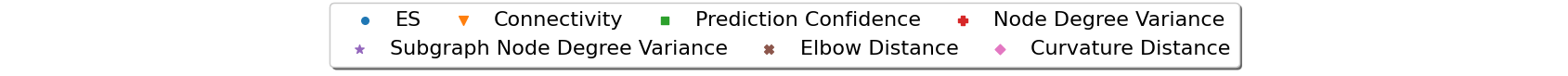}

            \caption{F1 score of each metric across varied trigger sizes, for attacks with random triggers. The random figures incorporate results from 544 attack configurations. The diversity of performance across individual metrics across all configurations validates the need to use metrics collectively rather than individually.}
            \label{fig:individual_metrics_triggersize}
        \end{figure*}

\subsection{Additional Ablation Studies}

        \subsubsection{Attack Strength vs. Individual Metric Performance}
        Fig. \ref{fig:individual_metrics_triggersize} illustrates the effectiveness of each individual metric in the context of varied trigger sizes and datasets in random attacks. 
        Once again, we observe the impact of trigger size and attack strength on the resulting capabilities of each metric: in most instances, for most metrics, performance improves as trigger size increases.

        A comparison of all five subfigures underscores the observations from Fig.~\ref{fig:boxplot} that the performances of individual metrics vary significantly between datasets; consider Connectivity, which performs worst among all seven metrics on PROTEINS, but best for IMDB-BINARY. The diversity of metric performance emphasizes the need to consider our metrics collectively.

        \vspace{+2mm}

        \subsubsection{Optimal NPMR at Varied Thresholds.}
        
        Table \ref{tab:percentiles_best_performance} presents the average F1 scores corresponding to the optimal NPMRs across various thresholding settings, with values in bold reflecting our default 25th/75th percentile threshold. At the default 25th/75th threshold, F1 scores range from 0.731 to 0.871. For all attack types, F1 scores begin a gradual decline at an upper-bound percentile of 80, reinforcing the suitability of the 75th percentile as a threshold for maintaining strong performance. However, these F1 scores generally do not see a sharp drop until an upper-bound percentile of 90; the stability of these metrics across a broad range of thresholds highlights the robustness of our detection method to variations in thresholding.

            \begin{table*}[h]
                \centering
                \footnotesize
                \begin{tabular}{c|ccccccccccc}
                    \toprule
                    {} & & \multicolumn{8}{c}{Percentile} \\
                     & 50 & 55 & 60 & 65 & 70 & 75 & 80 & 85 & 90 & 95 & 100 \\
                    \hline
                    Random & 0.769 & 0.788 & 0.802 & 0.803 & 0.813 & \textbf{0.795} & 0.771 & 0.741 & 0.771 & 0.621 & 0.389 \\
                    Clean Label & 0.933 & 0.929 & 0.920 & 0.909 & 0.894 & \textbf{0.871} & 0.838 & 0.786 & 0.838 & 0.535 & 0.246 \\
                    TRAP & 0.768 & 0.772 & 0.771 & 0.770 & 0.762 & \textbf{0.790} & 0.662 & 0.574 & 0.438 & 0.277 & 0.090 \\
                    LIN & 0.708 & 0.727 & 0.747 & 0.768 & 0.786 & \textbf{0.797} & 0.786 & 0.744 & 0.639 & 0.415 & 0.075 \\
                    Motif       & 0.800 & 0.792 & 0.772 & 0.762 & 0.762 & \textbf{0.744} & 0.728 & 0.687 & 0.728 & 0.547 & 0.384 \\
                    Adaptive    & 0.746 & 0.749 & 0.751 & 0.748 & 0.740 & \textbf{0.731} & 0.705 & 0.698 & 0.705 & 0.559 & 0.291 \\

                    \bottomrule
                \end{tabular}
                \caption{Average F1 score at optimal NPMR, across varied clean value thresholds. The values in bold correspond to our default 25th/75th percentile threshold.} 
                \label{tab:percentiles_best_performance}
            \end{table*}

        \begin{figure*}[!t]
            \centering
            \includegraphics[width=1\linewidth,trim=0 0 0 0, clip]{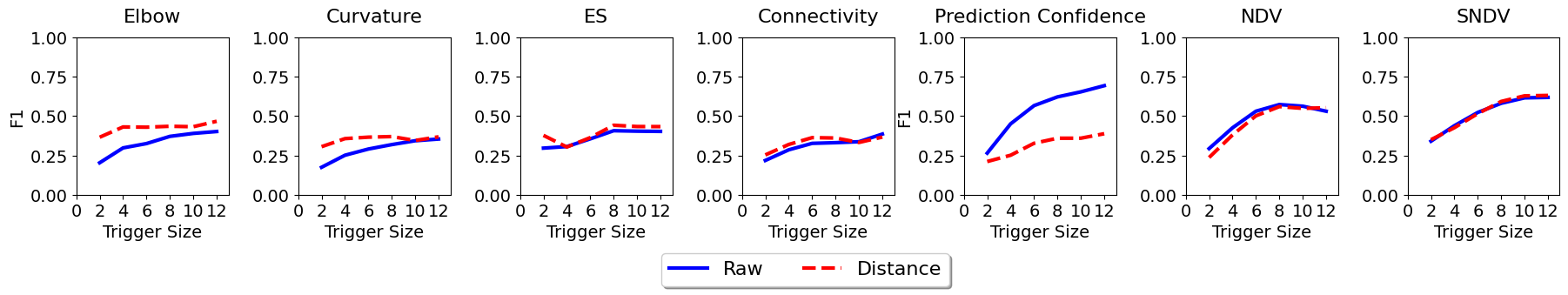}    
            \caption{Raw and distance metric performance across trigger sizes in random attacks.}
            \label{fig:raw_vs_dist}
        \end{figure*}

    \subsection{Sensitivity Analyses}
    
        \subsubsection{Effect of Trigger Size on Raw and Distance Metrics}
        
        Fig.~\ref{fig:raw_vs_dist} compares the relationships between raw data and various distance metrics across different trigger sizes in random attacks. The insights gleaned from these comparisons shed light on the effectiveness of selecting distance-based metrics for Curvature and Elbow. The figure shows that the distance metrics for Curvature and Elbow consistently outperform their raw data counterparts, particularly when dealing with smaller trigger sizes. This is because subtle triggers do not strongly influence the explainer loss curve, causing the curve to closely resemble that of clean samples. In these cases, the expected behavior of the raw metrics—whether they should increase or decrease relative to clean data—can reverse, making raw metrics unreliable. By normalizing raw metric values to their distance from the clean validation distribution (Equation~\eqref{distance}), we reduce dependence on specific directional trends, focusing instead on the magnitude of deviation from clean behavior. This approach addresses the inconsistencies in raw metrics for loss curve analysis, offering improved performance in many cases where small triggers might otherwise go undetected.

    \subsubsection{Performance with Limited Clean Validation Data}
    Our method uses clean validation data, a common practice in backdoor defense methods~\cite{Pal2024BackdoorSU, Li2021AntiBackdoorLT, Xian2023AUD, Li2023ReconstructiveNP}. Establishing a clean validation baseline requires only a small number of samples. For example, while we use 50 clean validation samples by default, Table~\ref{table:clean_validation} demonstrates that comparable F1 detection scores can be achieved with just 20 clean validation samples.

        \begin{table}[!t]
            \begin{tabular}{l|llllllll}
                \toprule
                {} & \multicolumn{7}{c}{NPMR}\\
                Dataset & 1 & 2 & 3 & 4 & 5 & 6 & 7\\
                \midrule
                All     & 0.71 & 0.76 & 0.74 & 0.59 & 0.36 & 0.10 & 0.01 \\
                MUTAG   & 0.58 & 0.76 & 0.91 & 0.94 & 0.77 & 0.25 & 0.03 \\
                AIDS    & 0.76 & 0.79 & 0.74 & 0.53 & 0.25 & 0.05 & 0.00 \\
                PROTEINS & 0.76 & 0.82 & 0.82 & 0.67 & 0.41 & 0.12 & 0.00 \\
                DBLP    & 0.72 & 0.71 & 0.63 & 0.43 & 0.22 & 0.08 & 0.02 \\
                IMDB-BIN. & 0.68 & 0.70 & 0.64 & 0.50 & 0.30 & 0.09 & 0.02 \\
                IMDB-MULT. & 0.76 & 0.81 & 0.84 & 0.79 & 0.66 & 0.50 & 0.23 \\

                \bottomrule
            \end{tabular}
        \caption{F1 detection scores against random attack when limited to 20 clean validation samples.}
        \label{table:clean_validation}
        \vspace{-6mm}
        \end{table}


\bibliographystyle{IEEEtran}
\bibliography{IEEEtran}

\newpage


 




\vfill

\end{document}